\documentclass{article}
\usepackage{moreverb}
\usepackage{multirow}
\usepackage{array}
\usepackage{graphicx}
\usepackage{fix-cm}
\usepackage{times}
\usepackage{dirtytalk}
\usepackage{textcomp}
\usepackage{eurosym}
\usepackage{hyperref}
\usepackage{nomencl,etoolbox,ragged2e,siunitx}
\usepackage{enumitem}
\usepackage{rotating}
\usepackage{lscape}
\usepackage{tabularx}
\usepackage{amsmath}
\usepackage{subfigure}
\usepackage{todonotes}

\newcommand\BibTeX{{\rmfamily B\kern-.05em \textsc{i\kern-.025em b}\kern-.08em
T\kern-.1667em\lower.7ex\hbox{E}\kern-.125emX}}

\begin{document}

\begin{center}
\title{Tracking Skin Colour and Wrinkle Changes During Cosmetic Product Trials Using Smartphone Images}  
\vspace {1cm}
{Alan F. Smeaton$^{1,*}$, Swathikiran Srungavarapu$^1$, \\
Cyril Messaraa$^2$ and Claire Tansey$^2$}

\vspace{1cm}

{($^1$) Insight Centre for Data Analytics},\\
{Dublin City University},\\
{Glasnevin, Dublin 9}, {Ireland.} \\ 
\vspace{0.5cm}
{($^2$) Oriderm R\&D, \\
Oriflame, \\
Bray, Co Wicklow, Ireland}\\ 
\vspace{1cm}
Corresponding author:  $^*$ Alan F. Smeaton\\
Insight Centre for Data Analytics\\ 
Dublin City University\\
Glasnevin, Dublin 9, Ireland. \\
{alan.smeaton@dcu.ie}\\
\end{center}
\date{} 
\abstract{
\noindent
{\bf Background:} To explore how the efficacy of product trials for skin cosmetics can be improved through the use of con\-sumer-level images taken by volunteers using a conventional smartphone.  
\\
{\bf Materials and Methods:} 12 women aged 30 to 60 years participated in a product trial and had close-up images of the cheek and temple regions of their faces taken with a high-resolution  
Antera 3D CS camera at the start and end of a 4-week  period.  
Additionally, they each had ``selfies'' of the same regions of their faces taken regularly throughout the trial period.  
Automatic image analysis to identify changes in skin colour used three kinds of  colour normalisation and analysis for wrinkle composition 
identified edges and calculated their magnitude.
\\
{\bf Results:}  
Images taken at the start and end of the trial acted as baseline ground truth for normalisation of smartphone images and showed large changes in both colour and wrinkle magnitude during the trial for many volunteers.
\\
{\bf Conclusions:} Results demonstrate that regular use of selfie smartphone images within trial periods can add value to interpretation of the efficacy of the trial.
}
\maketitle

\footnotetext{\textbf{Abbreviations:} SIFT, Scale-invariant feature transform; CLAHE, Contrast-limited adaptive histogram equalisation}

\section{Introduction}\label{sec1}

Cosmetic companies  invest huge resources each year on researching new products that are specific to  consumer requirements. The claimed efficacy of these products needs to be substantiated and documented in the Product Information File, as specified by Article 11 of Regulation (EC) No 1223/2009. Clinical trials on volunteers are one of the possible approaches to generate relevant data with regards to the efficacy of cosmetic products. Conventionally such product trials begin by recruiting volunteers and measuring some baseline characteristics of their appearance in a lab setting before having them use the new product, or a placebo, in their everyday lives for a period of time, that can range from minutes (e.g. make-up, instant effect) to a number of weeks. At the end of the trial period, the volunteers return to the laboratory for a re-test of their appearance, allowing for a before-and-after comparison, with and without the product being evaluated.

A major problem with this approach is that testers do not know what their subjects have been doing during the trial period because there is no possibilities to monitor their compliance to the protocol. With skincare as the leading category amongst cosmetic products, if subjects have been sunbathing, or spending time in cold and harsh outdoor conditions which present challenging conditions for skin products then not having control over, or even knowing, this variable  reduces the usefulness of such product trials.

This paper explores the capabilities of a consumer smartphone (Huawei P20 PRO) in capturing  information on facial characteristics during cosmetic product trials.  For this study,  volunteers took part in  initial and end-of-trial assessments of their skin characteristics  and received skincare products to use during a 4-week trial period.  Volunteers' faces were imaged regularly during the trial  using the smartphone. Images were analysed by normalising and extracting  facial features thus tracking changes in  skin characteristics {\it during} the trial period. The work addresses if these intermediate images  add value to  before-and-after lab assessments as part of product trials.

\section{Materials and Methods}


\subsection{Subjects and Equipment}

We recruited 12 female volunteers with ages ranging from 30 to 60 years for this  study.  Volunteers attended an initial and an end-of-trial facial skin assessment in a laboratory environment and as many of the  intermediate smartphone image capture sessions as they could.
Each volunteer was given a randomly-generated id for anonymity purposes.  Ethical approval for this project  was granted by the University's School of Computing Research Ethics Committee.

The Huawei P20 PRO smartphone \cite{phoneref1} 
is co-engineered with Leica  and operates with Artificial Intelligence (AI) to guarantee the highest outcomes in photography are always achieved. The phone comes with a dual-lens camera, one at 12 MP (RGB, f/1.8 aperture) and the other at 20 MP (Monochrome, f/1.6 aperture) which support contrast focus, phase focus, laser focus, and deep focus. Its front camera is a single-lens camera of 24MP with an f/2.0 aperture and supports a fixed focal length. 2D Photos were taken of our subjects with the camera at a resolution of $5120 \times 3840$ pixels with f/1.8 aperture. 

The Antera 3D camera \cite{Linming_2017,Messaraa2018WrinkleAR} offers a powerful and easy-to-use technology for aesthetic medicine. It captures close up images of the skin using an enclosed light hood which means it operates under controlled lighting. It can be used to measure skin wrinkles, texture, scars, the colour of the skin, redness, and pigmentation \cite{Messaraa2018WrinkleAR}.  It offers a complete overview of the skin, and was used in the laboratory setting at the start and at the end of the trial period in the work reported here.

\subsection{Protocol}

We captured facial image data during a cosmetic product trial using the Antera 3D camera on the first and last day of a 4-week trial period. 
As we are interested in facial skin changes during the trial, we also captured facial images using the smartphone during the trial period including the first and last day.

On the first day of the trial, volunteers were relaxed in a controlled environment before the left temple and cheek areas of their faces were imaged using the Antera 3D and also the smartphone camera. Following that, each volunteer received an anti-ageing product kit and usage instructions. The product kit  comprised a cleanser,  toner, eye cream, serum, and day and night creams. 

During the trial period, cheek and temple images were recorded for each volunteer up to 3 times per week over a 4 week period, using the Huawei P20 PRO camera with the highest resolution. 
The colour values in an image are dependent on the varying lighting conditions, location, environment, and other factors.
Because the smartphone images were taken under different lighting conditions during the trial period,
to normalise  colour values in  the images and  determine the  true colour of the skin, we  used colour cards \cite{colourcardref1} composed of standard colours when taking the images.
Volunteers  held the colour card against their face, below the nose level for the image of their temple and across the eye for the image of their cheek.

\begin{figure}[htb]
    \centering
        \includegraphics[height=6cm]{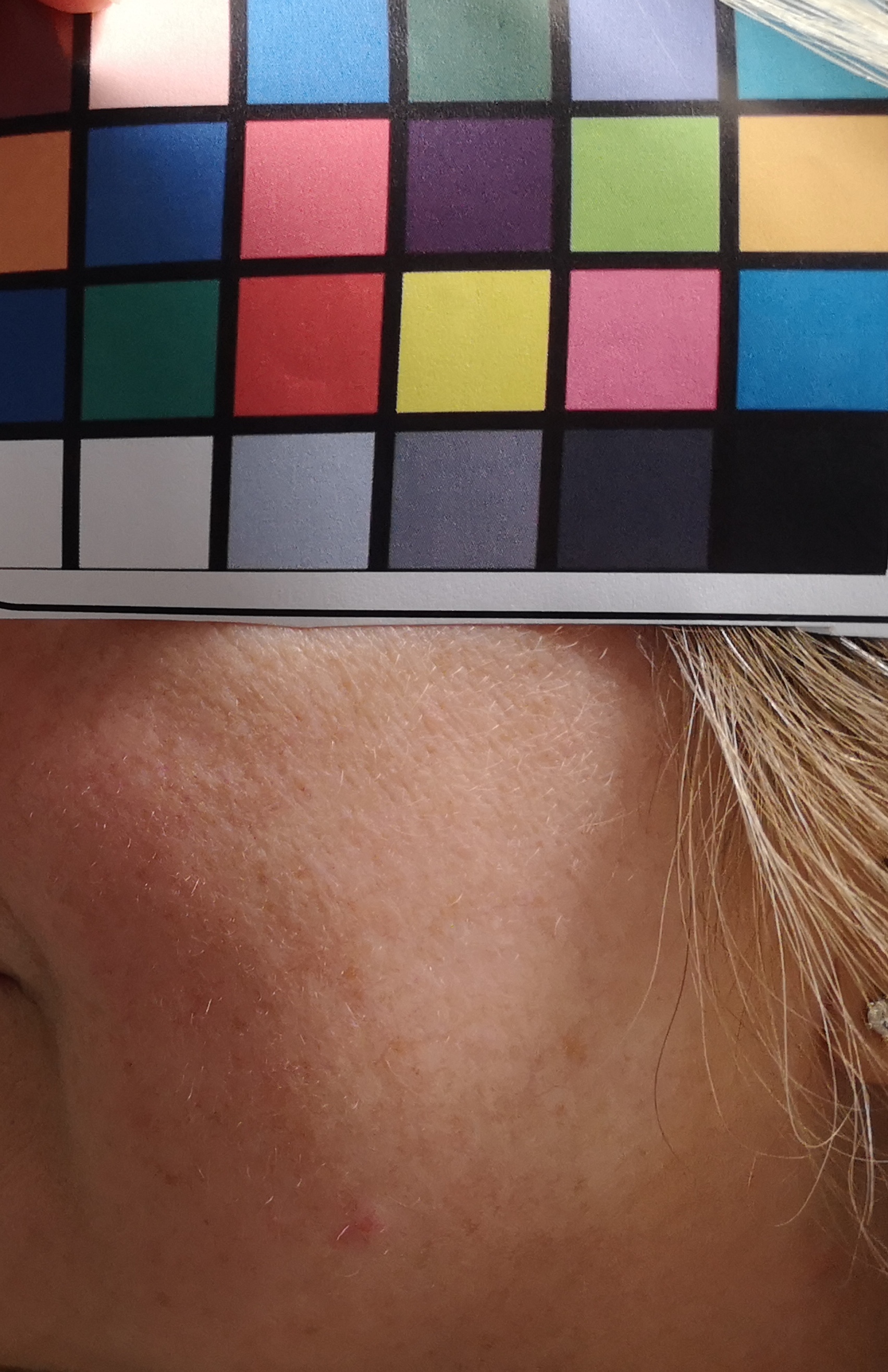}
        \hspace{0.05\textwidth}
        \includegraphics[height=6cm]{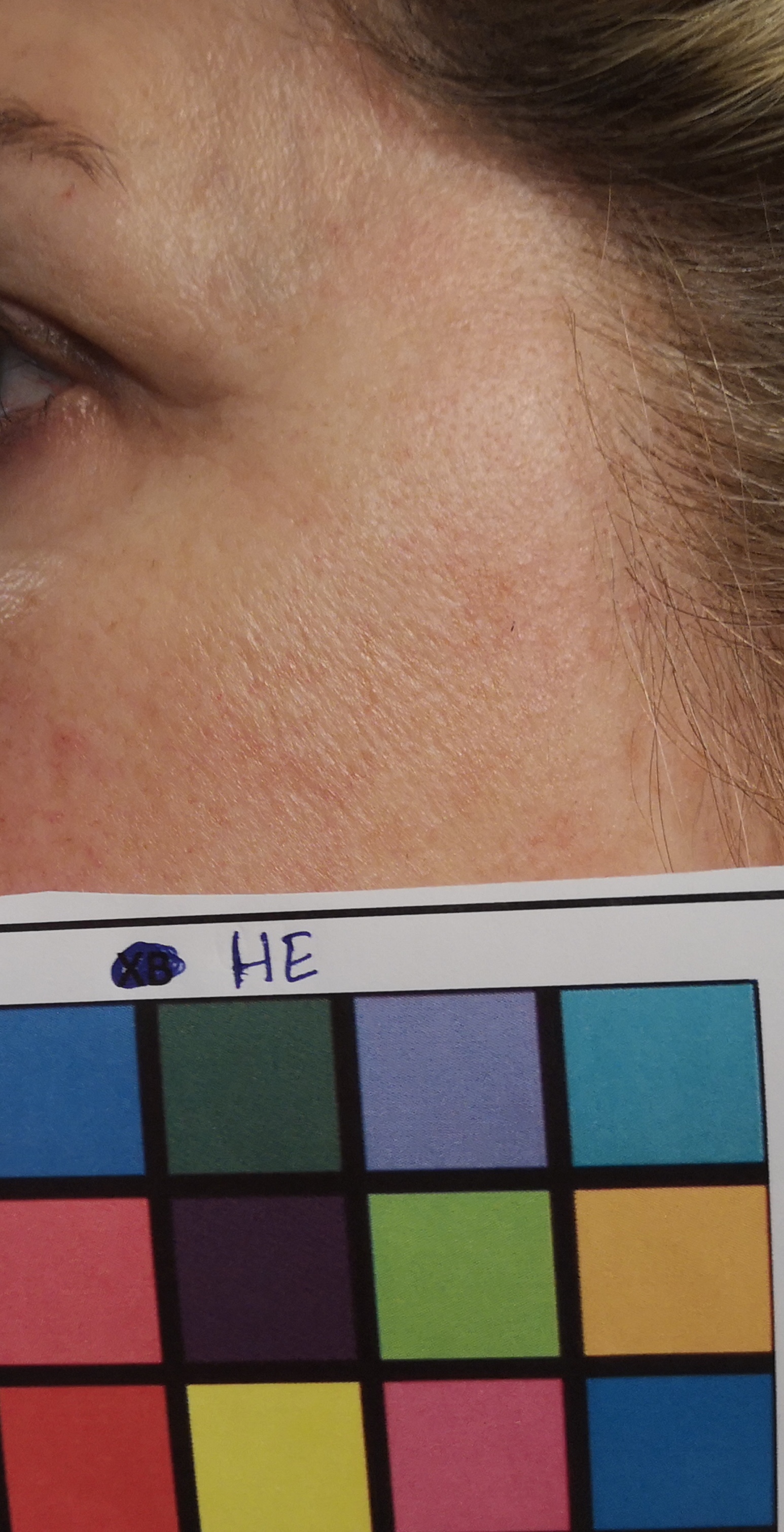}
    \caption{Cheek and Temple areas captured using Huawei P20 PRO smartphone with colour card shown}
    \label{fig:SmartphoneDataCollection}
    \label{fig:cheekArea}
\end{figure}

During the trial period, volunteers were imaged  at an indoor location on the University campus. 
The cheek and temple areas for two volunteers respectively, are shown in Figure~\ref{fig:cheekArea}.

\begin{figure*}[htp]
\centering
\includegraphics[width=0.9\textwidth, height = 5cm]{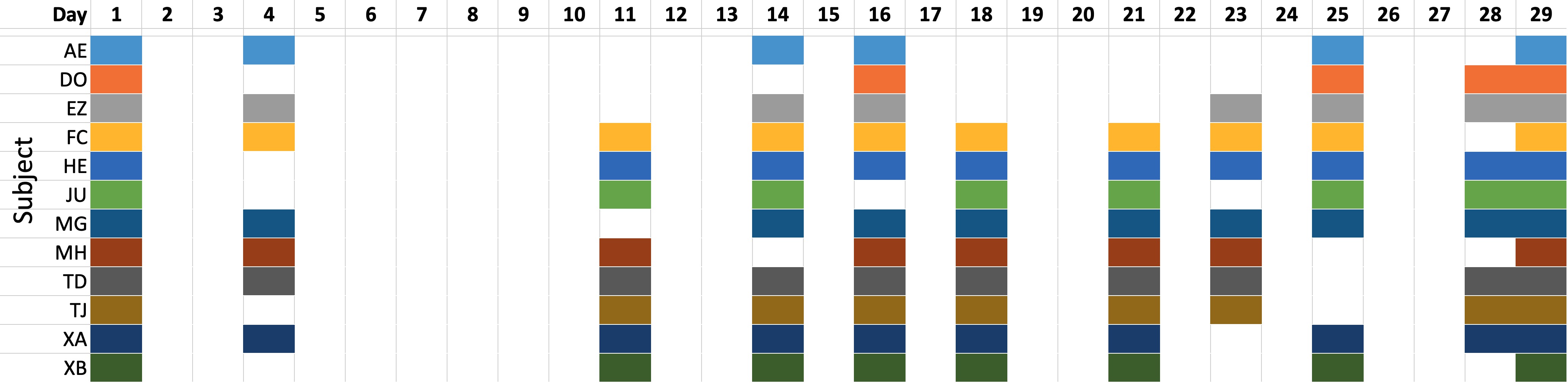}
\caption{Subject participation for within-trial imaging}
\label{fig:ParticipantsList}
\end{figure*}

On the last day of the trial period, a similar procedure of image capture to the initial day was repeated using the Antera 3D and the  smartphone camera. 
All 12 volunteers attended both the initial and final imaging sessions and volunteers also attended between 5 and all of the 10 scheduled within-trial imaging sessions (mean=8.5, std=1.68).
The full attendance record for volunteers is shown in Figure~\ref{fig:ParticipantsList}.  This completed our data collection.

\subsection{Image Processing for Skin Colour}
\label{subsec:normalisation}


The distribution of colour values in an image depends on the illumination, which varies widely under real-world conditions \cite{colourNorm}.  We explored  two known colour adjustment and transformation algorithms along with an approach based on  normalisation using colour cards which can be applied within an image.

\subsubsection{\textbf{Histogram Equalisation}}
Histogram equalisation  distributes the initial illumination intensity values in an image  uniformly, nearly equal to the image’s histogram \cite{finlayson2005illuminant}. The algorithm follows a series of monotonic, non-linear mappings to re-assign the intensity values of pixels in the input image such that the output image contains a flat histogram of intensities \cite{histNorm}.  We used the equalizeHist() function implemented in the OpenCV software library \cite{bradski2008learning}. However, this function cannot be applied directly to images in the RGB or LAB colour space therefore we converted the images into YUV colour space, where Y is the luminance component and UV are the colour components. We then then equalised the Y component alone. Later, the equalised images were converted back into RGB colour space.  Figure~\ref{fig:normImages} shows a sample image and its histogram equalised image.

\subsubsection{\textbf{CLAHE (Contrast Limited Adaptive Histogram Equalisation)}}
Histogram equalisation  considers  global image contrast during the image transformation process.  CLAHE is an adaptive method that computes several histograms  corresponding to  distinct sections in an image and then transforms these sections for redistribution of the lightness values of the image \cite{claheNorm}. It is therefore suitable for improving  local contrast and enhancing the definition of edges in  regions of an image where  the size of the regions is user-defined. In this work we used regions of $4 \times 4$ pixels during the CLAHE transformations. Figure~\ref{fig:normImages}  shows a transformation result using CLAHE.

\par
\subsubsection{\textbf{Image Normalisation Using Colour Cards}}

\begin{figure}[htp]
\centering
\includegraphics[width=0.35\textwidth]{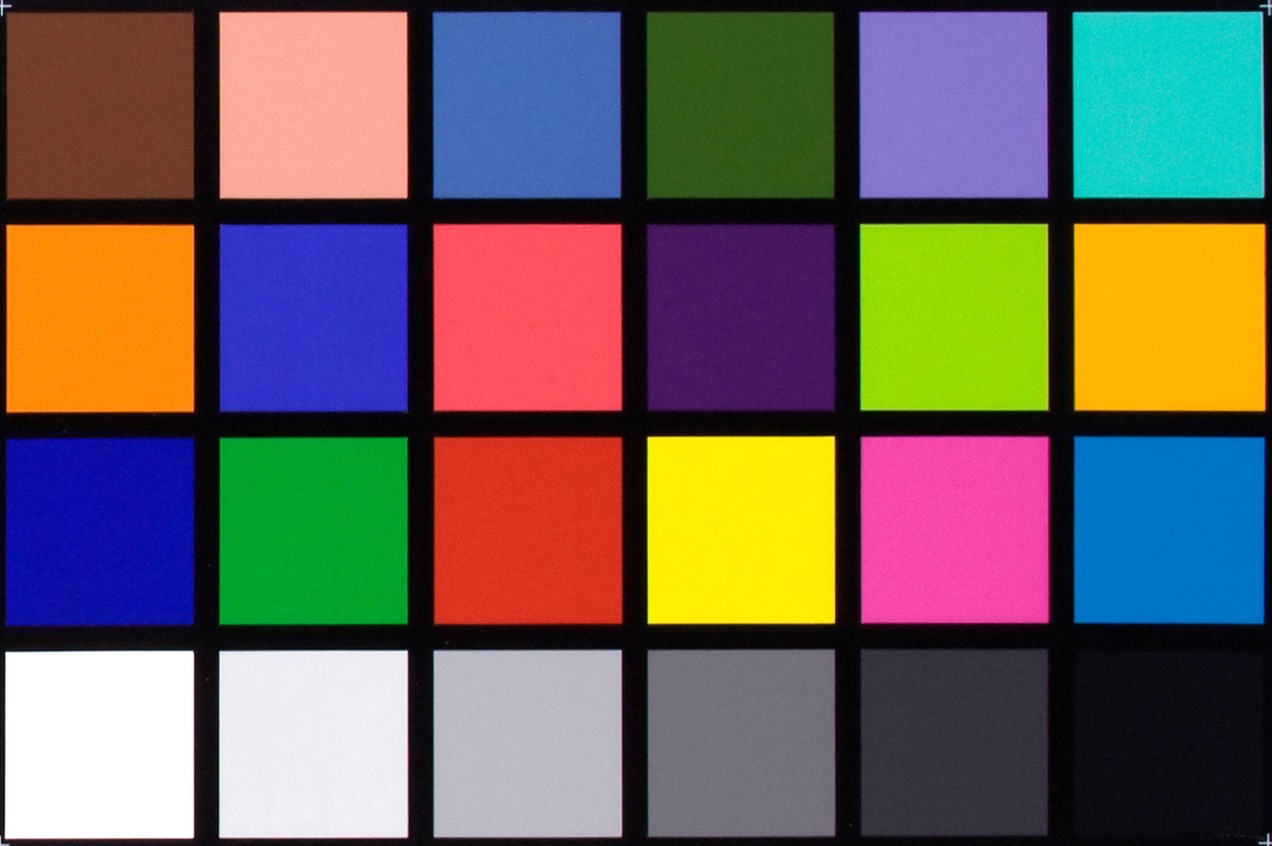}
\caption{Colour card used by volunteers}
\label{fig:ccard}
\end{figure}

Printed colour cards such as the one shown in Figure~\ref{fig:ccard} were used during  smartphone image capture. The aim  is to measure the differences between the known true  colour values from the cards and the  values observed in the captured images and to adjust the images accordingly. Ideally, we try to adjust luminance and colours in the image to account for the real-world conditions in which the images are taken.

\noindent
If L, A, B are the true values of Black/White, Blue/Red, Yellow/Green respectively for the colour purple at row 2 and column 4 of the colour card in Figure~\ref{fig:ccard} and, let P be the same area with the colour purple in a captured image I then  d(L) = L1,     d(A) = A-A1  and         d(B) = B-B1
where, L1 is the calculated mean value of Black/White for area P, 
A1 is the calculated mean value of Blue/Red in area P, and 
B1 is the calculated mean value of Yellow/Green in area P.  We let 
d() be the set consisting of the difference values which are applied to an image captured with a smartphone d is a subset of \{d(L), d(A), d(B)\}

\begin{figure}[htb]
    \centering
        \includegraphics[width=0.24\textwidth]{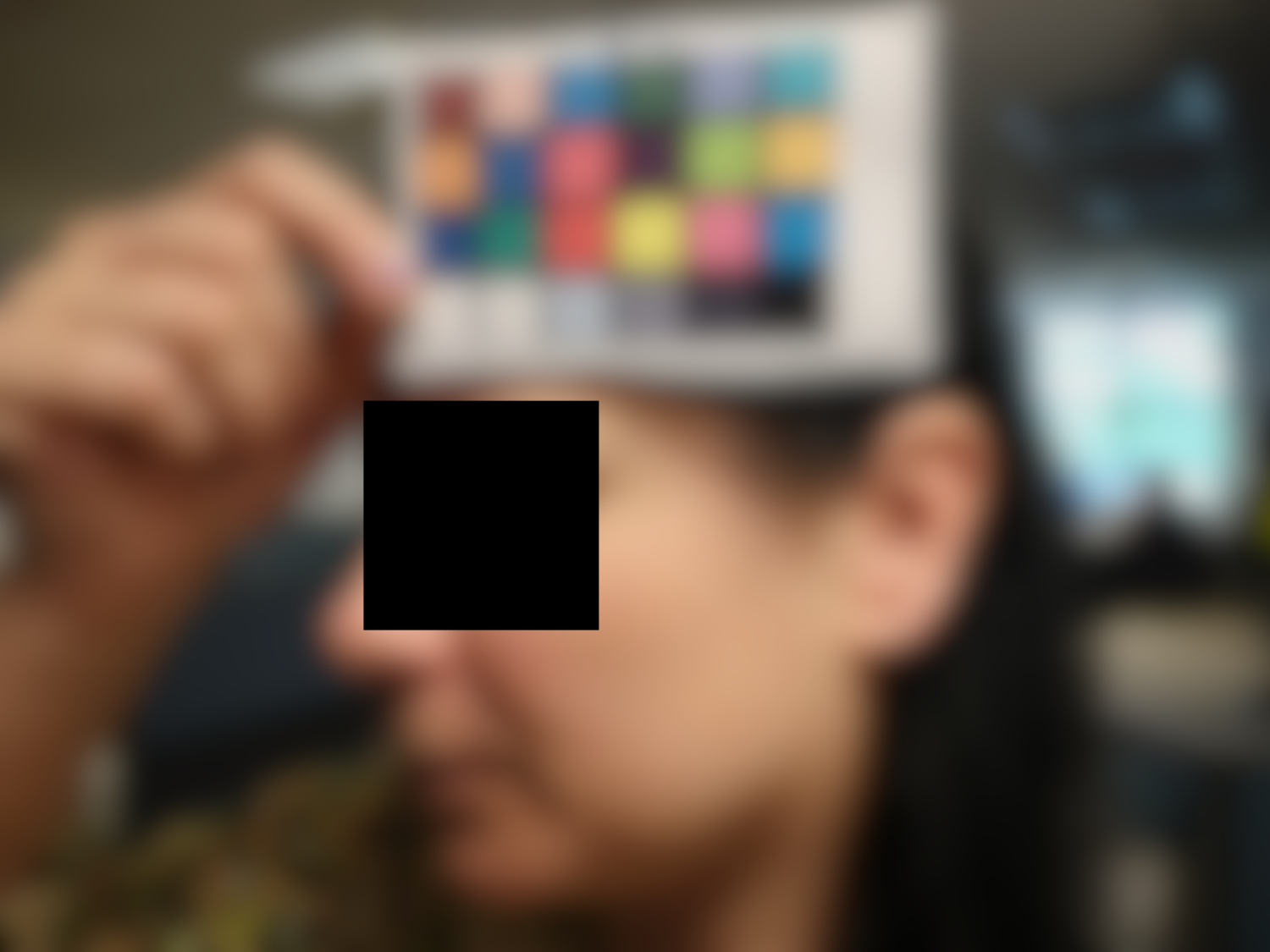}
        \includegraphics[width=0.24\textwidth]{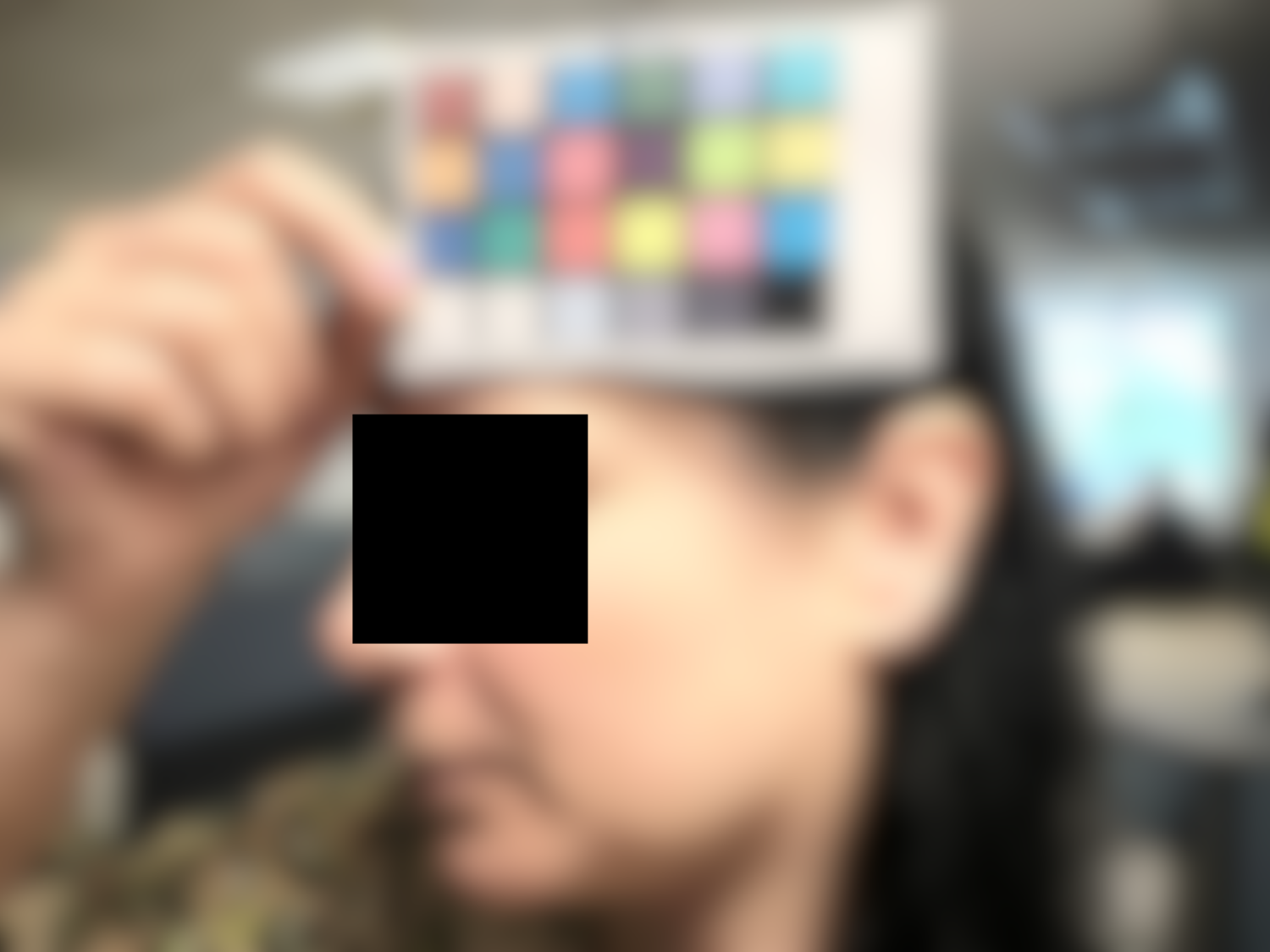}
        \includegraphics[width=0.24\textwidth]{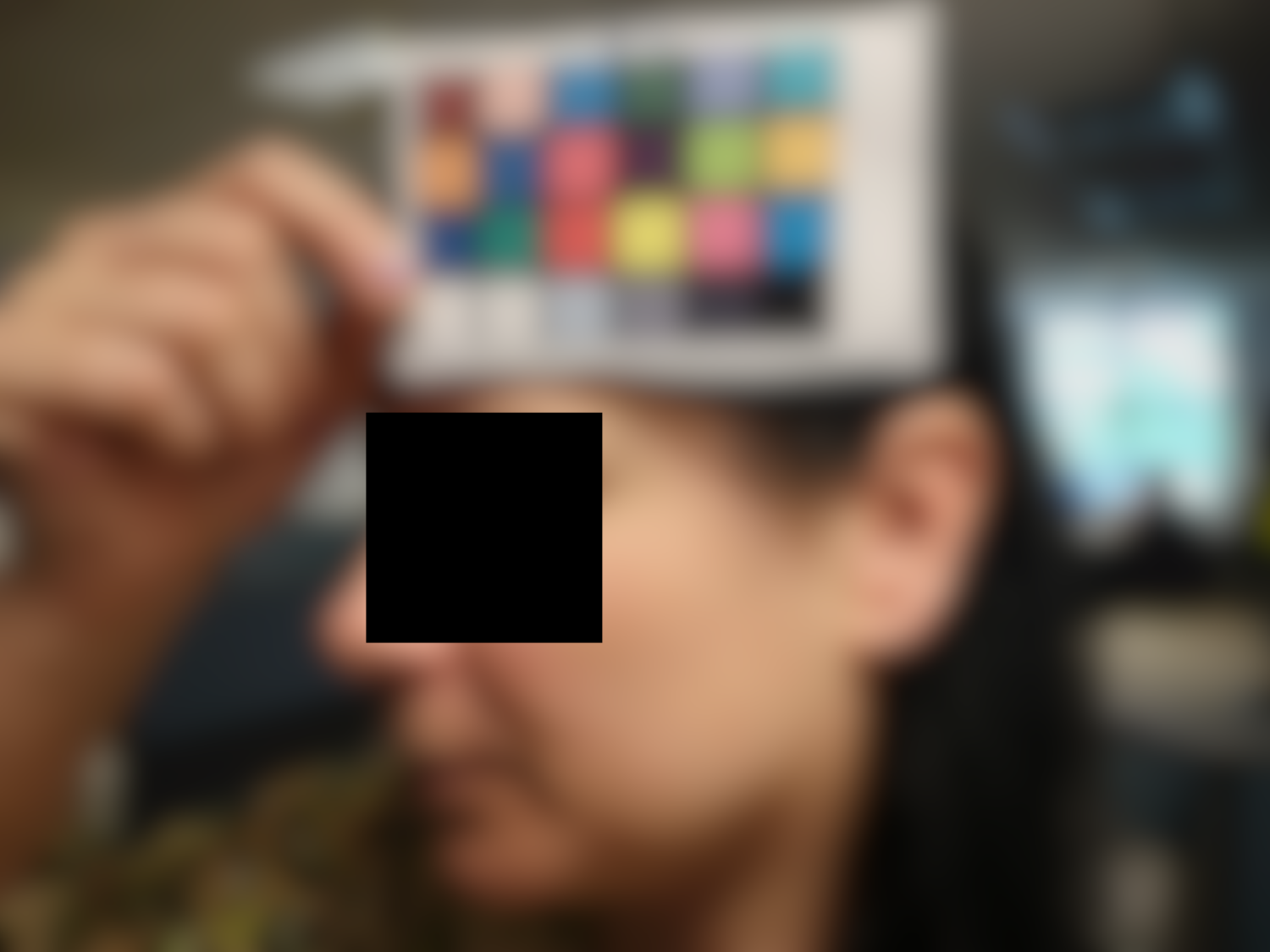}
        \includegraphics[width=0.24\textwidth]{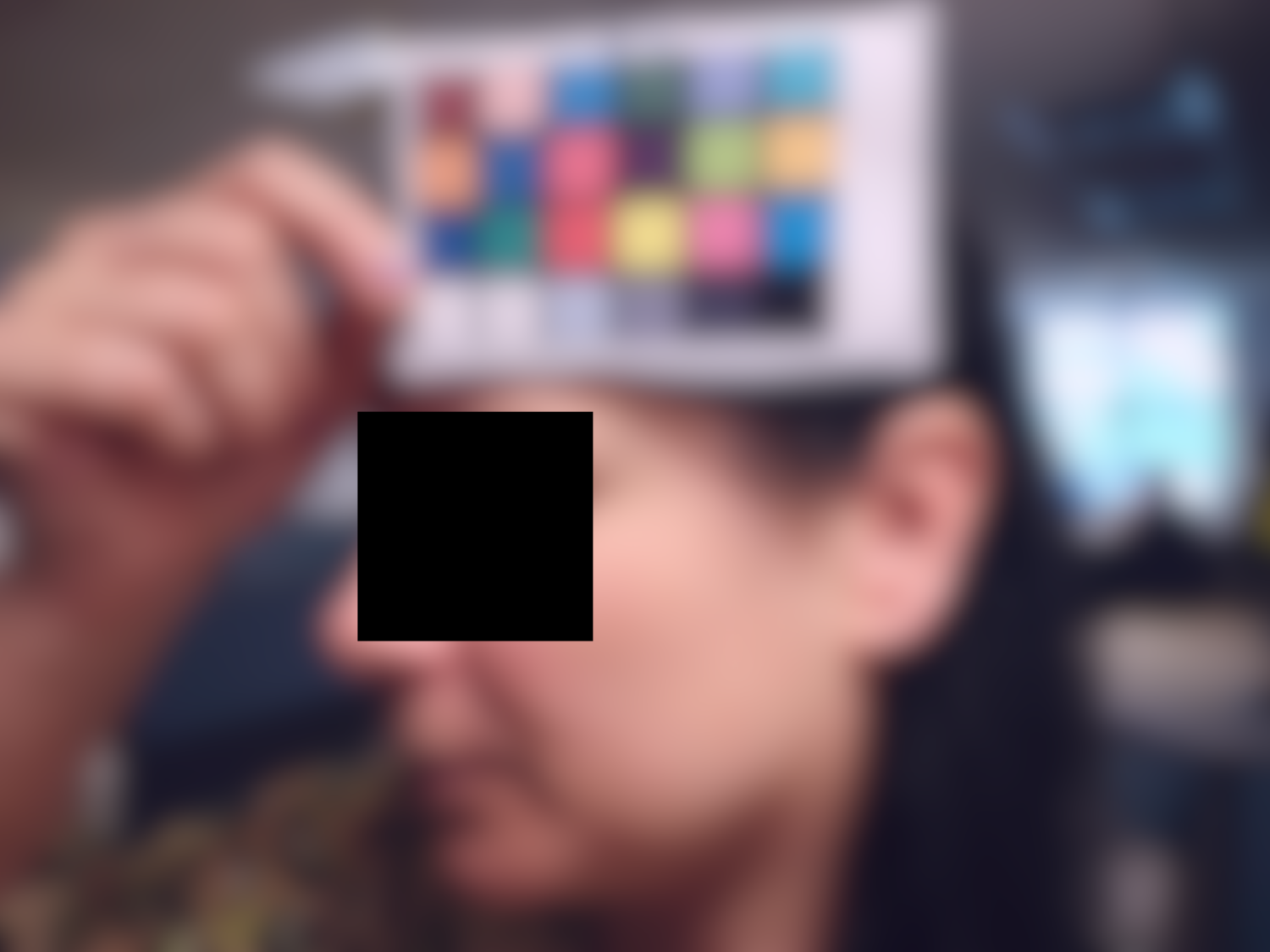}
    \caption{Normalising images showing the original,  histogram equalised, CLAHE and colour card normalisation. Note that  this and  subsequent images are blurred for volunteer anonymity.}
    \label{fig:normImages}
\end{figure}

Each image of a volunteer  captured using the smartphone with the colour card included in the image was transformed using all three normalisation techniques. Figure~\ref{fig:normImages}  illustrates that there can be clear  differences among the outputs of different techniques.

\subsubsection{\textbf{Estimating Skin Colour}}
Once  images are normalised for colour using each of the above methods, the next step is to calculate skin colour values using pixels in the cheek area. Normalised images are maintained in the LAB colour space rather than RGB because the LAB colour space is used by the Antera 3D camera. Every image pixel in the LAB colour space is described by the amount of black/white, green/red and yellow/blue respectively. The cheek area is used for calculating the average L, A, B values that represent the volunteer's skin colour from each image. These values define the amount of brightness, redness, and yellowness of the skin.

\subsection{Image Processing for Wrinkle Measurement}
\label{sec:FeatureExtractionWrinkle}

Feature detection is the process of automatically computing an abstraction of an image and making a local decision at every point in the image to determine if there is an image feature of a given type which exists at that point \cite{featureMatchRef}.  We explored methods  to automatically align  images of the same volunteer taken during the trial period so we could exactly match the skin areas, and then extract a common area of the face for wrinkle comparison across the set of images thus tracking wrinkle changes across the trial period.  In the end we used a semi-automated technique based on SIFT matching.

\begin{figure}[ht]
    \centering
        \includegraphics[height=4cm]{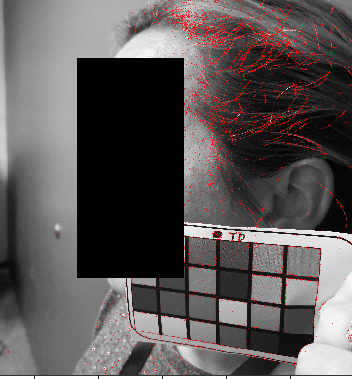}
        \includegraphics[height=4cm]{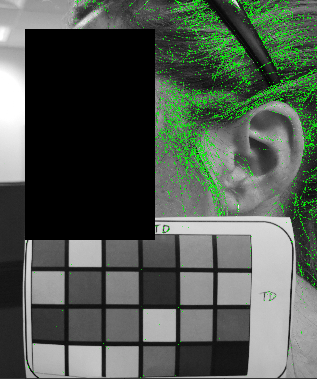}
        \includegraphics[height=4cm]{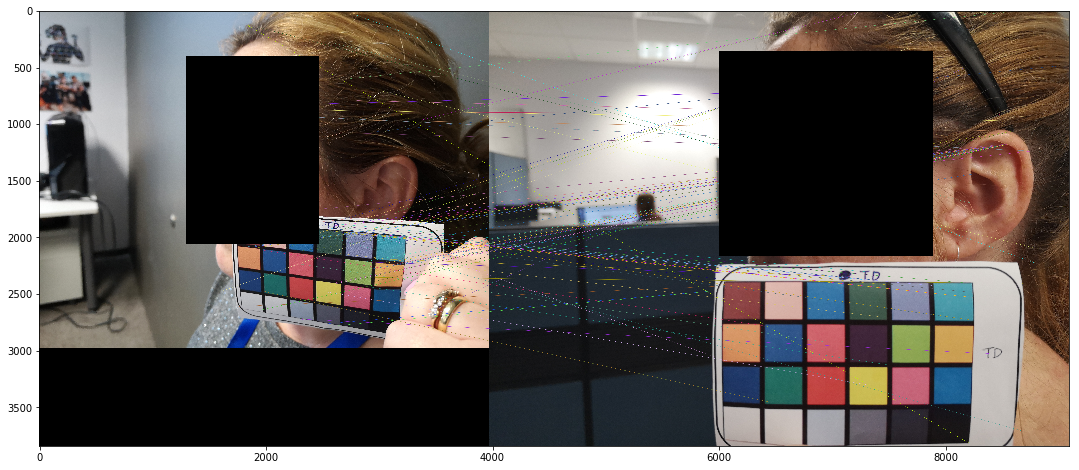}
    \caption{Key-points and matching using SIFT showing training image, a query image and SIFT-kNN matching.}
    \label{fig:SIFTTotal}
\end{figure}

Scale Invariant Feature Transform (SIFT) \cite{SIFT_matching} is a popular local feature descriptor which can be used to match similar or near-identical images. The general approach of this algorithm is to extract many key-points from  images and then  compute a local descriptor for each of these key-points. Image-to-image matching or scoring is then performed by matching each key-point descriptor in a query image against all descriptors extracted from a target image and summing the matches across point descriptors. Figure~\ref{fig:SIFTTotal} shows image matching based on SIFT features.

Following feature detection, we aligned  images of each volunteer and and detected the common wrinkle area  across  images taken during the trial period. The wrinkle area in our study is the cropped portion of the temple image consisting only of wrinkles.
%
Once this is completed,  images are processed  to measure wrinkle magnitude using edge detection, with the following steps.

Edge detection is less complex when working with grayscale images compared to other colour spaces therefore the gradient magnitude in this study is calculated using grayscale images \cite{gradientDetetction}. Gradient magnitude is a quantification of the edges in an image where a gradient operator simply produces large amplitudes in the output image indicating edges in the input image. Therefore, the more edges in the image the greater the amplification and vice-versa. We  apply the Laplacian method to compute the gradient magnitude \cite{mlsna2009gradient}. 



\begin{figure}[htb]
    \centering
        \includegraphics[width=0.23\textwidth]{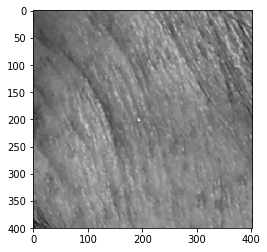}
        \includegraphics[width=0.23\textwidth]{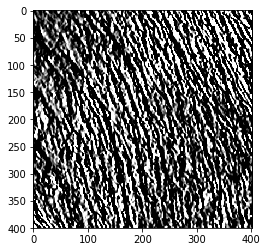}
        \includegraphics[width=0.23\textwidth]{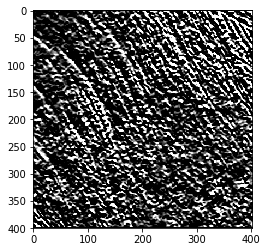}
        \includegraphics[width=0.23\textwidth]{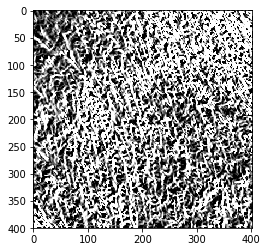}
        \label{fig:sobelCombinedImage}
    \caption{Gradient Edge detection stages showing grayscale, Sobel X, Sobel Y and Sobel Combined}
    \label{fig:gradientEdgeDetectionImage}
\end{figure}

The Sobel  edge detector  is a classic computer vision operator which uses a derivative approximation to find the edges in an image. The Sobel operator in a 2-dimensional image works in both directions, that is Sobel-X for left-to-right or horizontal scanning and Sobel-Y for top-to-bottom or vertical scanning for edge detection. Edges detected in both directions at a given point in an image can be combined using the logical OR operator and this is referred to as Sobel-combined. Figure~\ref{fig:gradientEdgeDetectionImage} illustrates the output of these operators on a sample image.

Using these  attributes of an image, we can estimate a measure for a volunteer's skin wrinkles as follows.

\begin{equation}
        W_G = S_G / I_G
\end{equation}
where,
$W_G$ is the Wrinkle Ratio of the Image G,
$S_G$ is the calculated mean of the Sobel Combined Image and
$I_G$ is the calculated mean of the Original Image G.

\subsection{Statistics}
The following sets of statistics were generated.
\begin{itemize} 
\item {\bf Antera 3D at baseline and after 4 weeks:}
Antera 3D parameters were analysed at baseline and following 4 weeks of treatment of the skin care treatment using paired t-test, or Wilcoxon matched pairs test if normality of the data was not observed with Shapiro-Wilk. Level for significance was set at $ \alpha = 0.05$.

\item {\bf Discrepancies between Smartphone parameters and Antera 3D parameters:}
Differences between Antera 3D  parameters and Smartphone parameters were evaluated by the mean of Mean Squared Error  (MSE)

\item {\bf Diurnal variation of skin parameters with Smartphone parameters:} 
Regular  evaluation of the parameters from the smartphone (colour and wrinkles) were plotted for each volunteer in order to visualise their individual diurnal variation.
\end{itemize}

\section{Results}
\label{sec:Results}

\noindent 
{\bf Antera 3D at baseline and after 4 weeks}

\noindent
Before we examine the per-participant results we will look at overall changes in colour and in wrinkles observed in results from the Antera 3D camera taken on the first and last days of the trial. Table~\ref{table:OriflameFinalResults} presents a summary. A significant reduction of wrinkle max depth and skin yellowness b* was observed following 4 weeks of using a skin care routine (paired t-test, $P < .05$).  Other parameters were not found to change significantly. 

\begin{table*}[htb]
\centering
 \begin{tabular}{|l||l|l|l|l|l|l|} \hline
 \multirow{2}{*}{\bf Parameters} & \multirow{2}{*}{\bf Colour (L)} & \multirow{2}{*}{\bf Colour (A)} & \multirow{2}{*}{\bf Colour (B)} & {\bf Wrinkle} & {\bf Wrinkle} & {\bf Wrinkle} \\
&&&& {\bf overall size} & {\bf depth} &{\bf max depth} \\ 
 \hline\hline
 \% variation & 0.8\% & 1.7\% & 4.0\% & 1.6\% & 2.0\% & 11.2\% \\ 
 \hline
 Significant & No & No & \textbf{Yes} & No & No & \textbf{Yes}\\
 \hline
 p value & 0.067 & 0.303 & \textbf{0.010} & 0.664 & 0.669 & \textbf{0.048} \\
 \hline
 Statistical Test & paired t-test & paired t-test & paired t-test & paired t-test & paired t-test & paired t-test \\ 
  \hline
\end{tabular}
\caption{Changes in colour and wrinkle assessment between start and end of trial using the Antera 3D camera}
\label{table:OriflameFinalResults}
\end{table*}

\begin{figure*}[htb]
    \centering
    \includegraphics[width=0.27\textwidth]{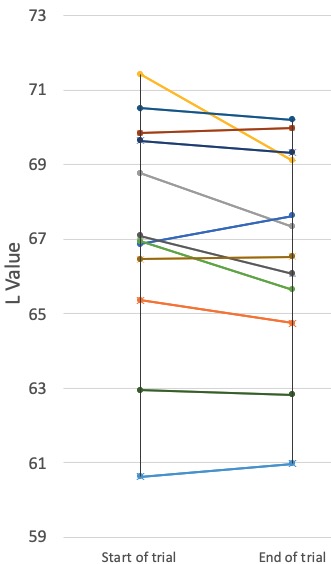}
    \hspace{0.5cm}
    \includegraphics[width=0.27\textwidth]{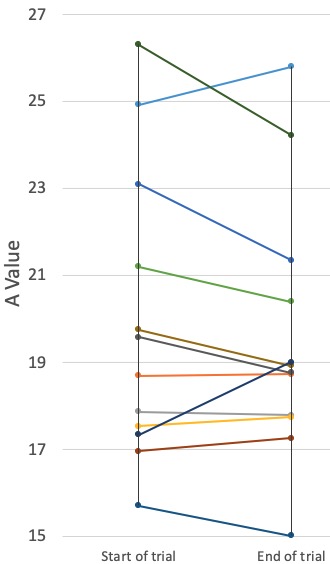}
    \hspace{0.5cm}
    \includegraphics[width=0.27\textwidth]{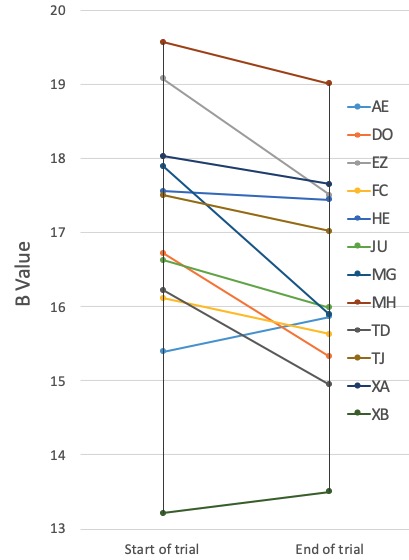}
    \caption{Skin colour analysis results for all volunteers at the start and end of the trial period using the Antera 3D camera.}
    \label{fig:lab-assessments-colour}
\end{figure*}

LAB colour values for  skin regions in an image are made up of an \textbf{L} value for describing the lighting, an \textbf{A} value for redness of the skin and a \textbf{B} value for describing the yellowness. 
We compared colour values from images taken with  the controlled lighting Antera 3D camera on the first  day of the trial with those taken on the last day on a per-volunteer basis, and the results are shown in Figure~\ref{fig:lab-assessments-colour}.
These  show that between the beginning and the end of the 4-week trial period, from among all volunteers for the A and B colour values there is a slight reduction for almost all volunteers except  JU, AE and HE. For the L values the overall trend is also slightly downward  except for FC and to a lesser extent EZ, TD and JU who have greater declines, while volunteer HE has the most noticeable increase. So there are lots of observed colour changes among volunteers but we do not know the nature of these changes  during the trial.
\\

\noindent 
{\bf Discrepancies between Smartphone parameters and Antera 3D parameters}

\noindent
We  examined the colour values for images taken using the smartphone which had already been normalised using each of the three within-image normalisation methods described in section \ref{subsec:normalisation}, and compared these against colour values from the Antera 3D camera.
Those differences were measured using Mean Squared Error (MSE) and Table~\ref{table:MSEtable} shows the MSE for each colour parameter and each colour normalisation technique with a range of  differences for different normalisation techniques.

\begin{table}[h]
\centering
 \begin{tabular}{|l|c|c|c|} 
 \hline
 {\bf Image Type} & {\bf MSE~(L)} & {\bf MSE~(A)} & {\bf MSE~(B)} \\ 
 \hline\hline
 Original & 19.25 & 42.05 & 22.51 \\ 
 \hline
 CLAHE & 61.75 & 45.29 & 21.74 \\
 \hline
 CC Normalised & 130.89 & 48.13 & 54.92 \\
 \hline
 Histogram & 255.52 & 99.16 & 14.18 \\
 \hline
\end{tabular}
\caption{MSE of normalisation methods}
\label{table:MSEtable}
\end{table}

We  calculated correlation between the colour parameters observed by the Antera 3D camera at the start of the trial, and the  colour parameters from the smartphone on day 1, normalised using three different approaches, and the results are shown in Table~\ref{tab:correls}.

\begin{table}[h]
\centering
 \begin{tabular}{|c|c|c|c|c|} 
 \hline
 {\bf Colour Parameter} &   {\bf ColorNorm} &  {\bf Histequal} &  {\bf CLAHE} &  {\bf Original} \\
 \hline \hline 
 L* & -0.159	&0.334	&0.628 &	0.719 \\
 A* & 0.237	&0.439	&0.860&	0.771 \\
 B* & -0.254	&-0.423	&-0.421&	-0.404 \\
  \hline
\end{tabular}
\caption{Correlation between Antera 3D and Smartphone skin colours}
\label{tab:correls}
\end{table}

What this shows is that colour normalisation techniques have varying levels of effectiveness and that the B element, corresponding to skin yellowness, is particularly difficult to replicate with a smartphone.  The A element corresponding to skin redness, which we are most interested in, correlates well with the Antera 3D when CLAHE normalisation is used with a correlation of 0.860.

\vspace{0.25cm}

\noindent 
{\bf Diurnal variation of skin parameters with Smartphone parameters}

\begin{figure*}[htb]
    \centering
    \includegraphics[width=0.33\textwidth]{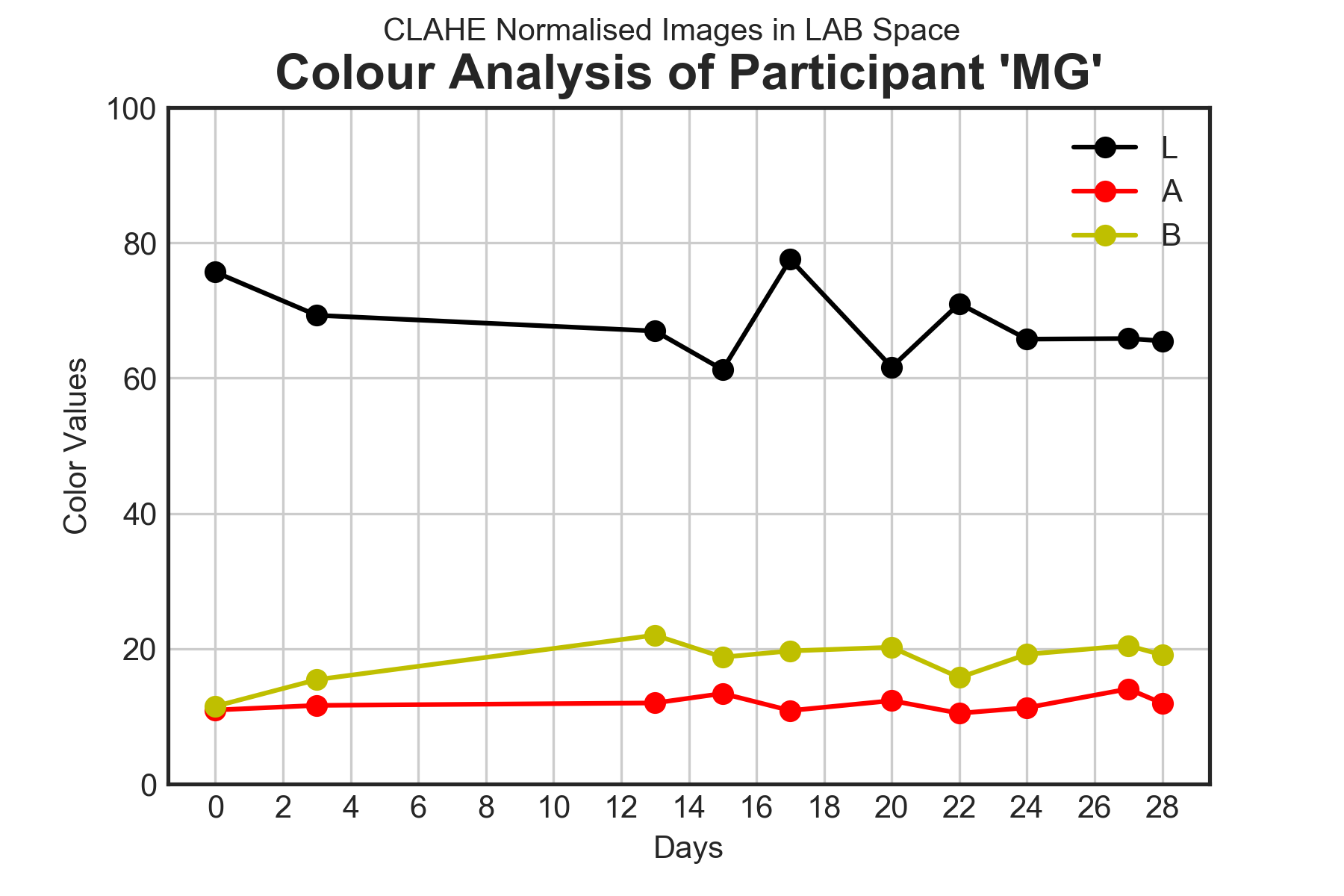}
    \vspace{0.2cm}
    \includegraphics[width=0.33\textwidth]{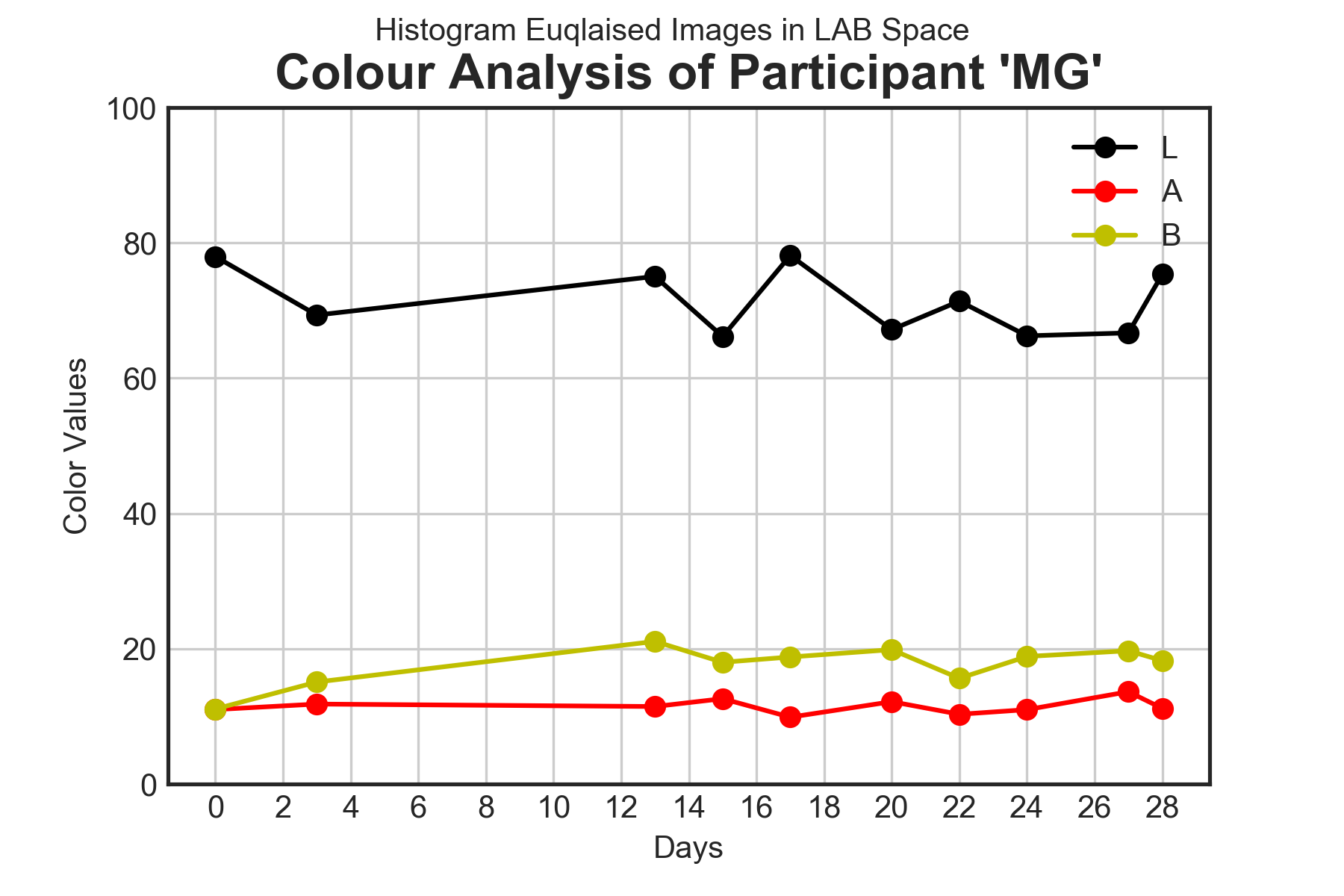}
    \vspace{0.2cm}
    \includegraphics[width=0.33\textwidth]{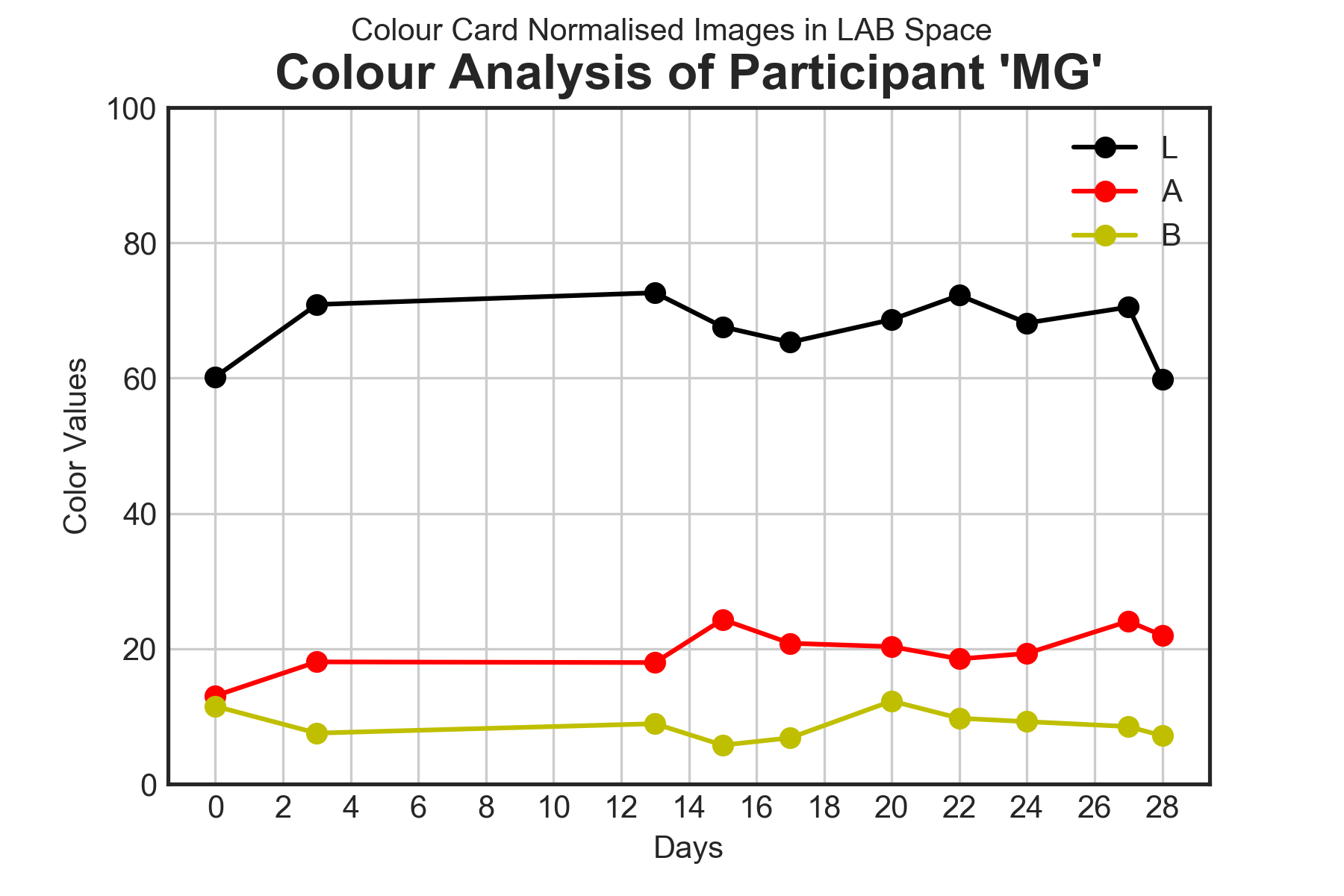}
    \vspace{0.2cm}
    \includegraphics[width=0.33\textwidth]{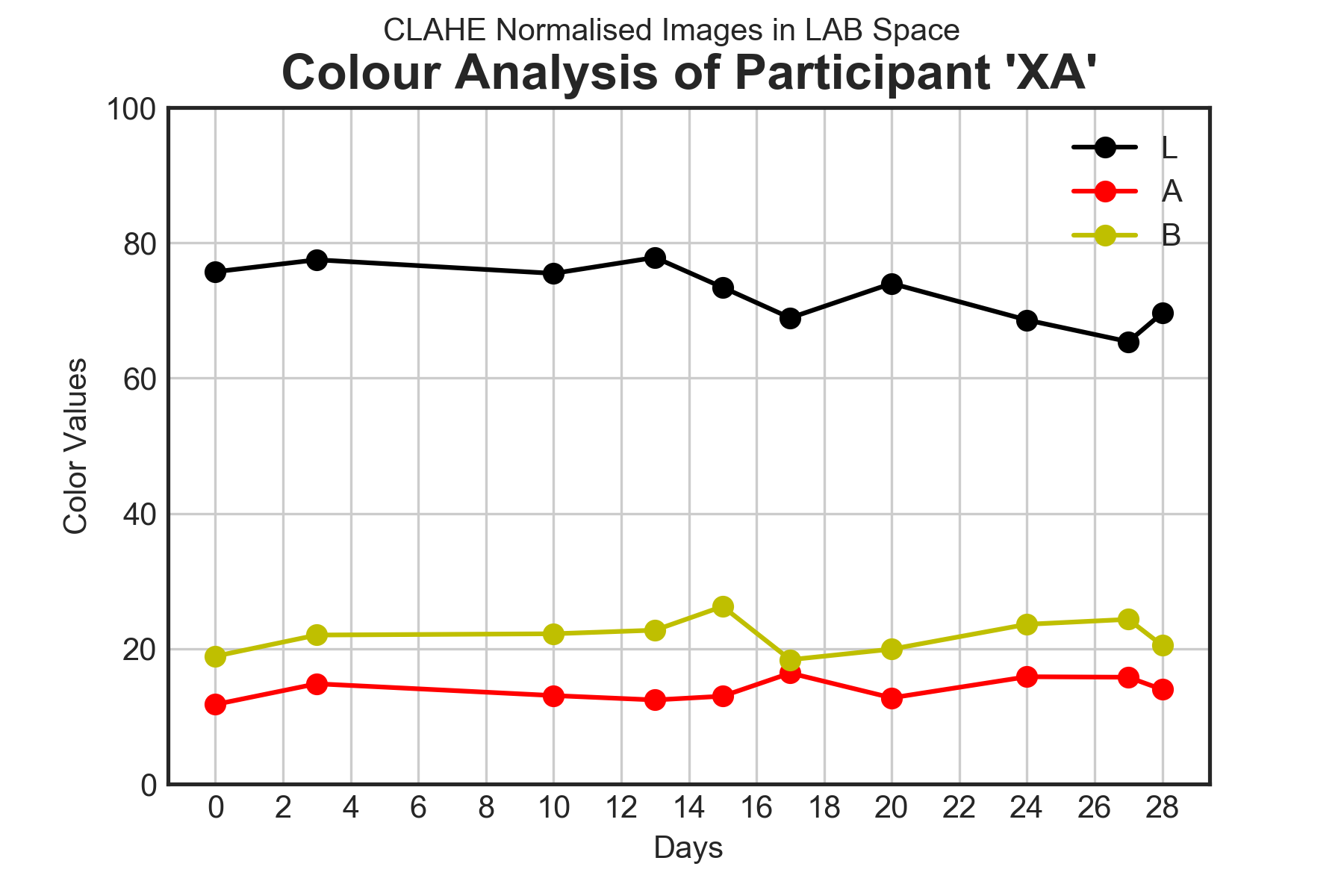}
    \vspace{0.2cm}
    \includegraphics[width=0.33\textwidth]{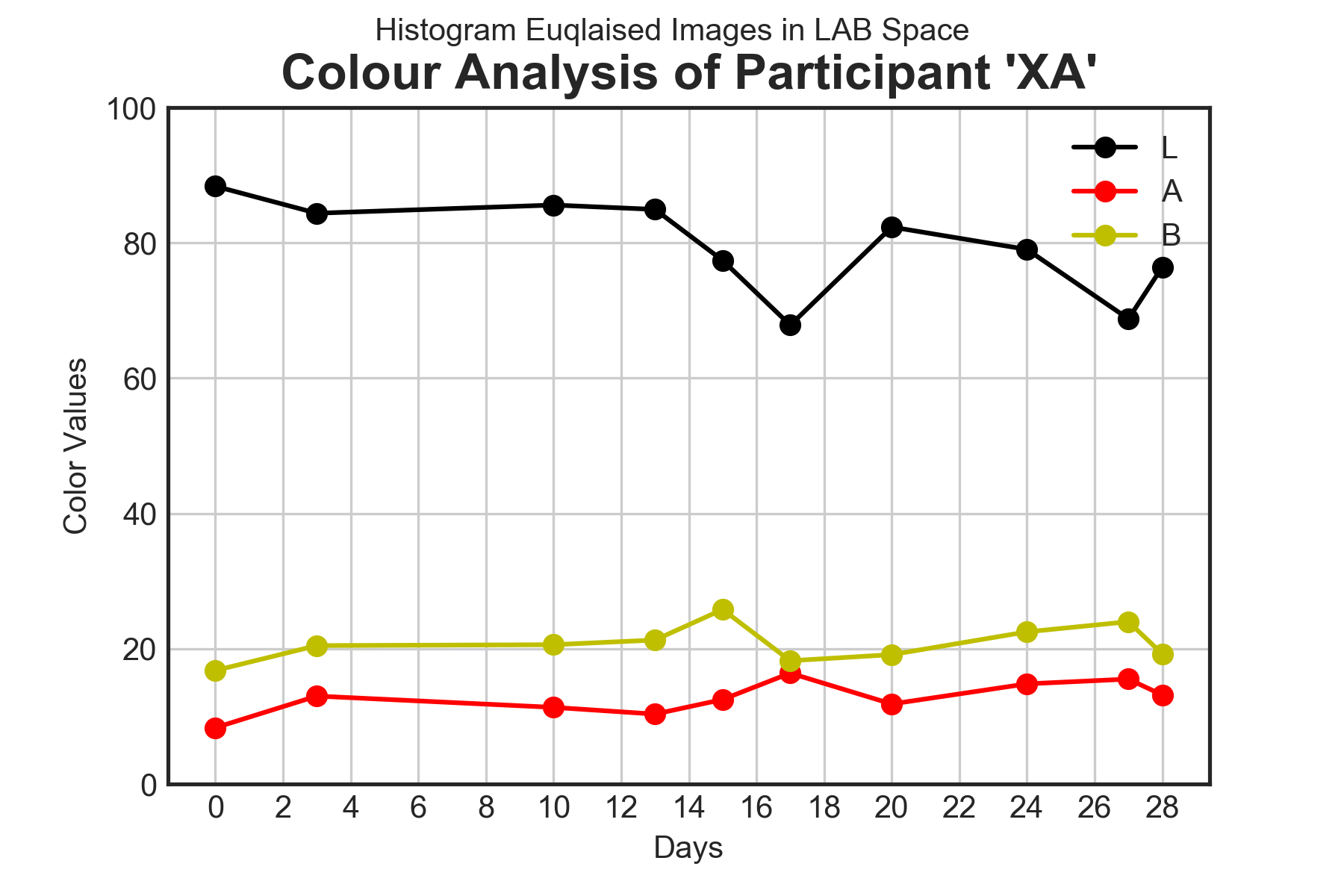}
    \vspace{0.2cm}
    \includegraphics[width=0.33\textwidth]{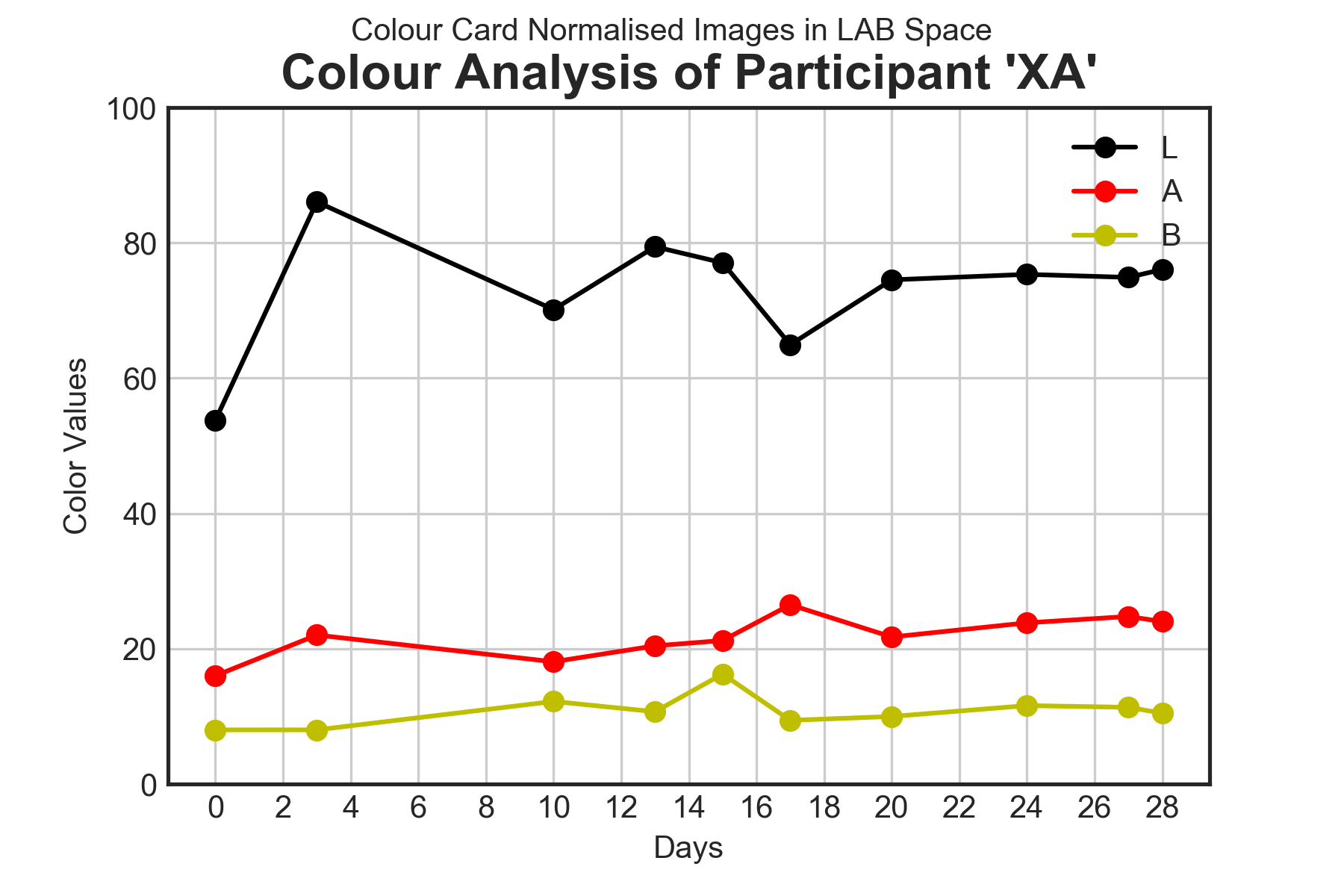}
    \vspace{0.2cm}
    \includegraphics[width=0.33\textwidth]{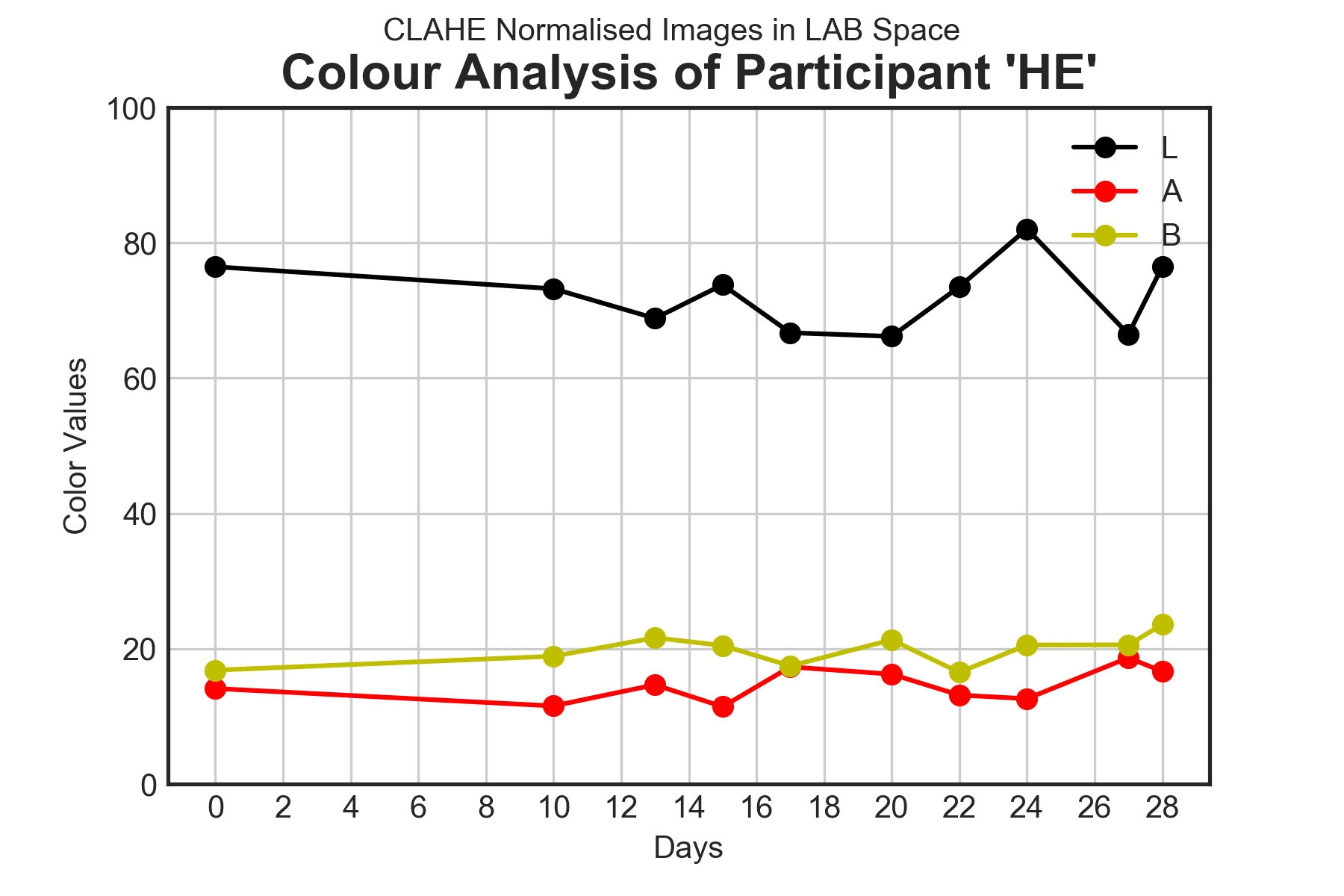}
    \vspace{0.2cm}
    \includegraphics[width=0.33\textwidth]{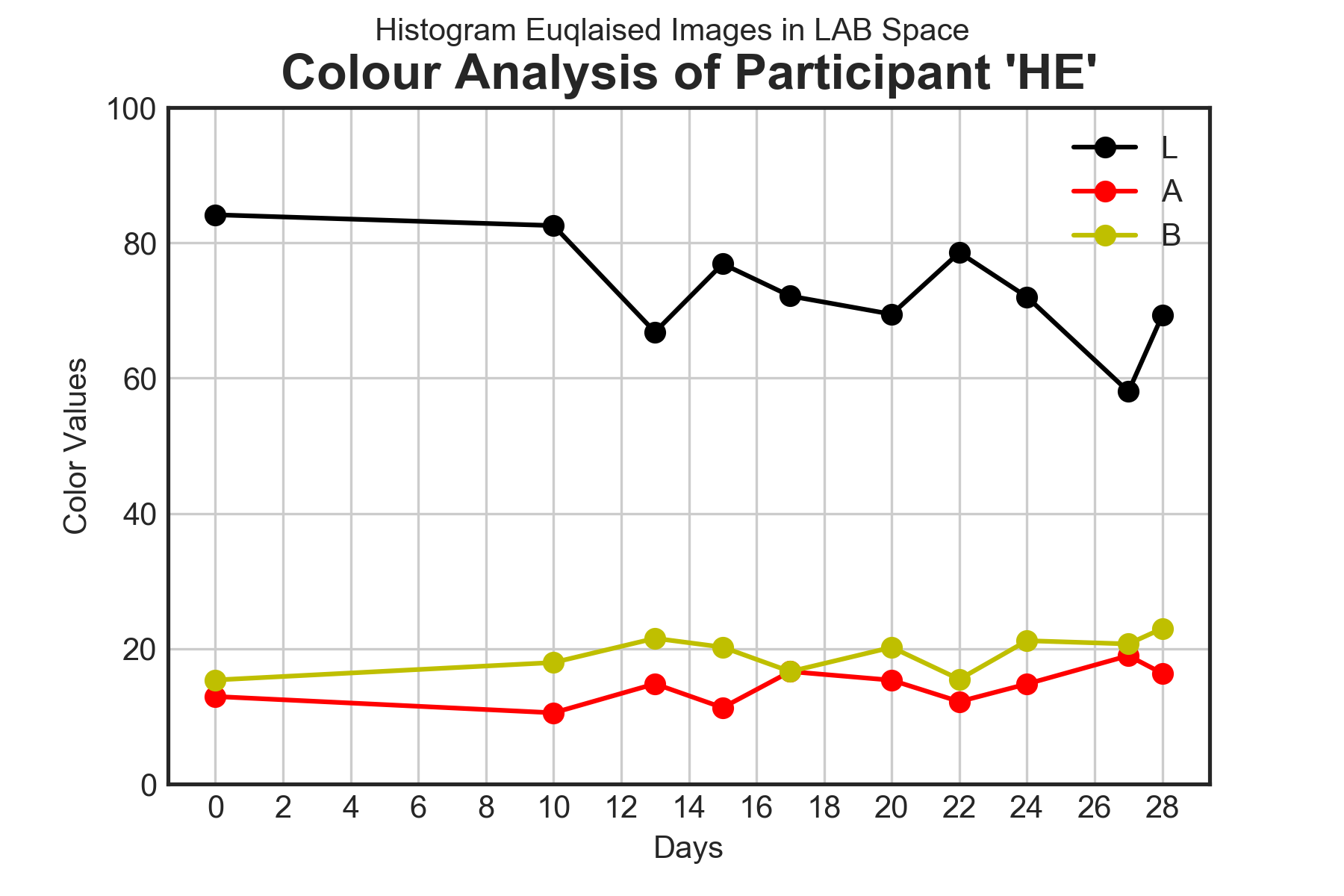}
    \vspace{0.2cm}
    \includegraphics[width=0.33\textwidth]{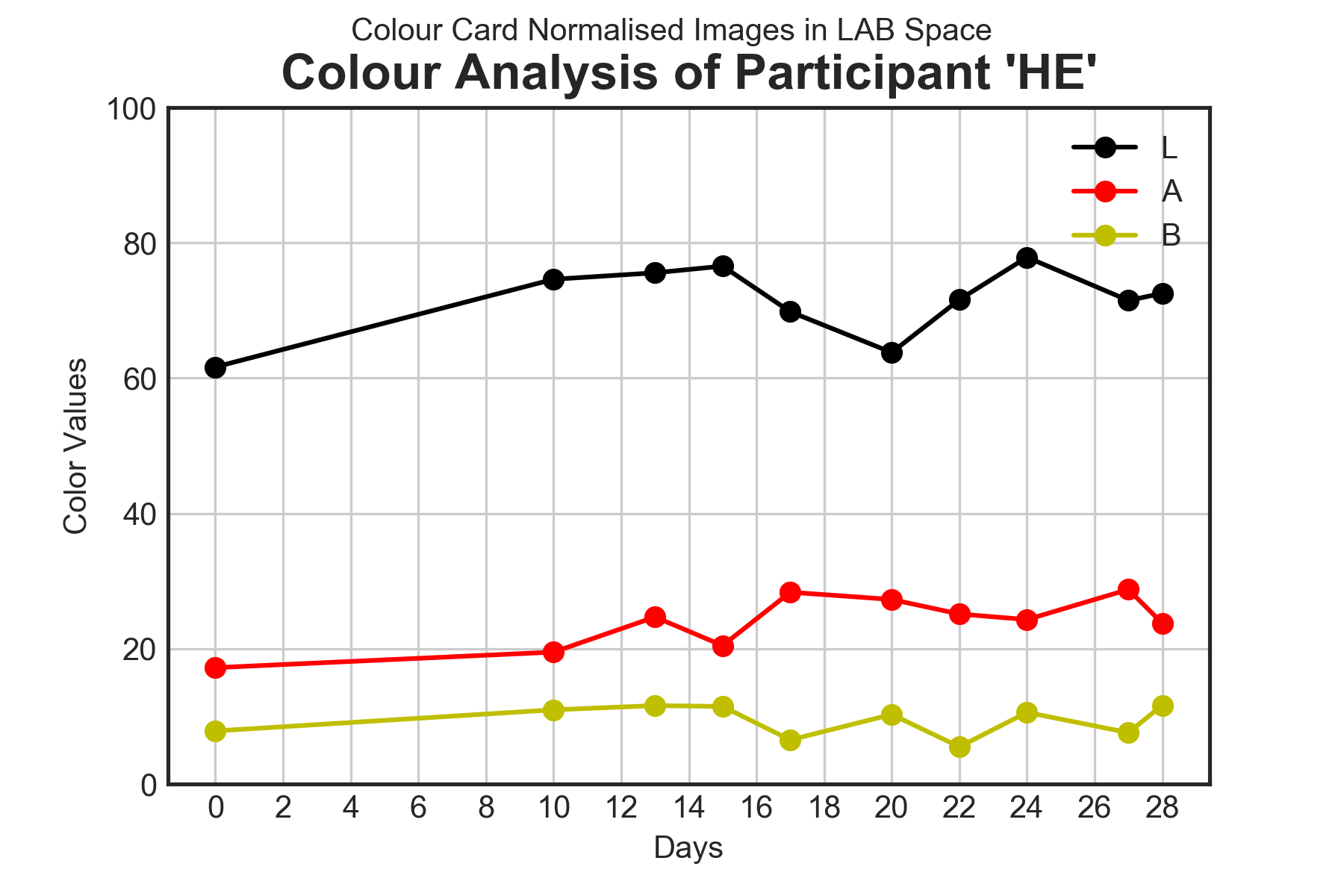}
    \vspace{0.2cm}
    \includegraphics[width=0.33\textwidth]{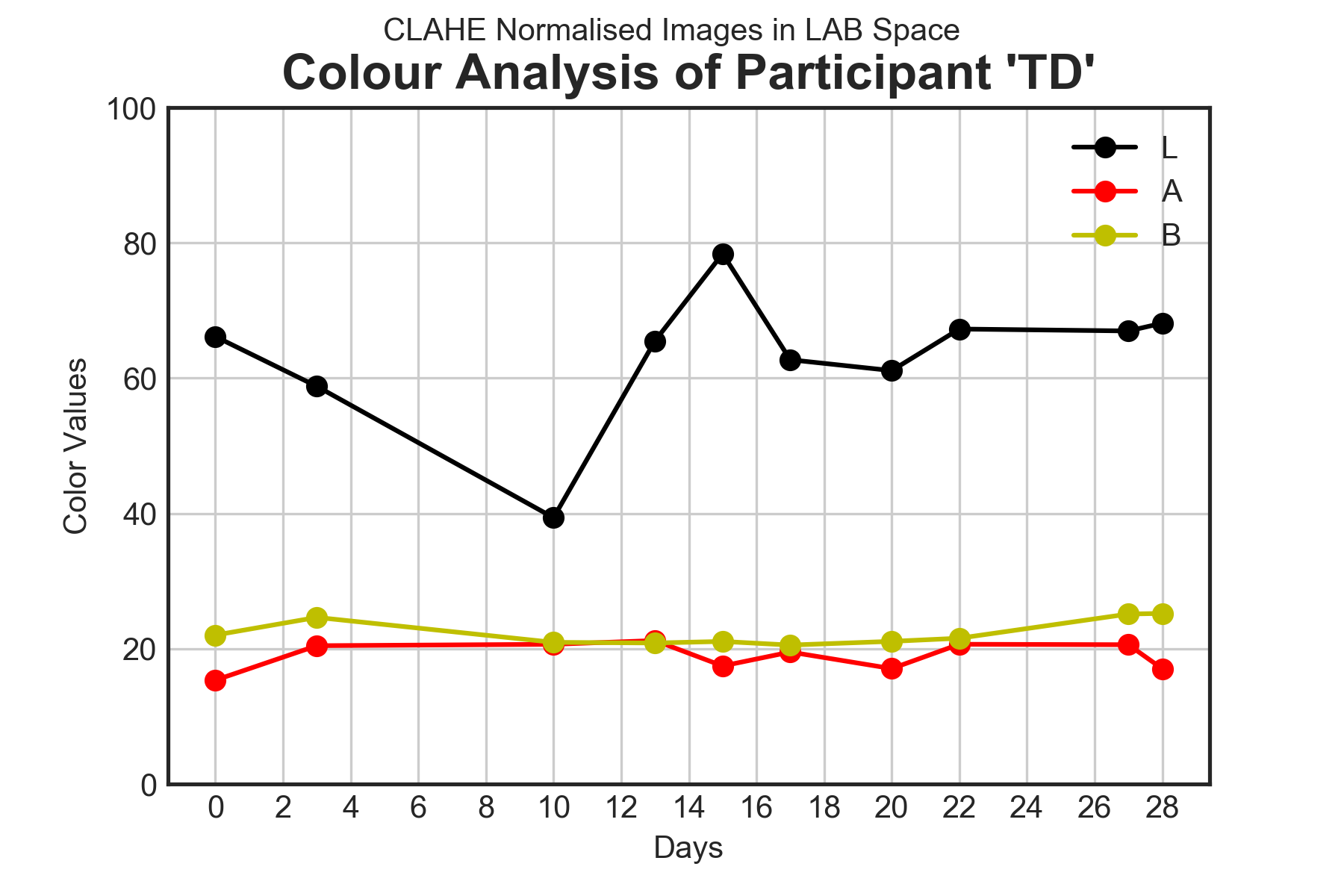}
     \vspace{0.2cm}
    \includegraphics[width=0.33\textwidth]{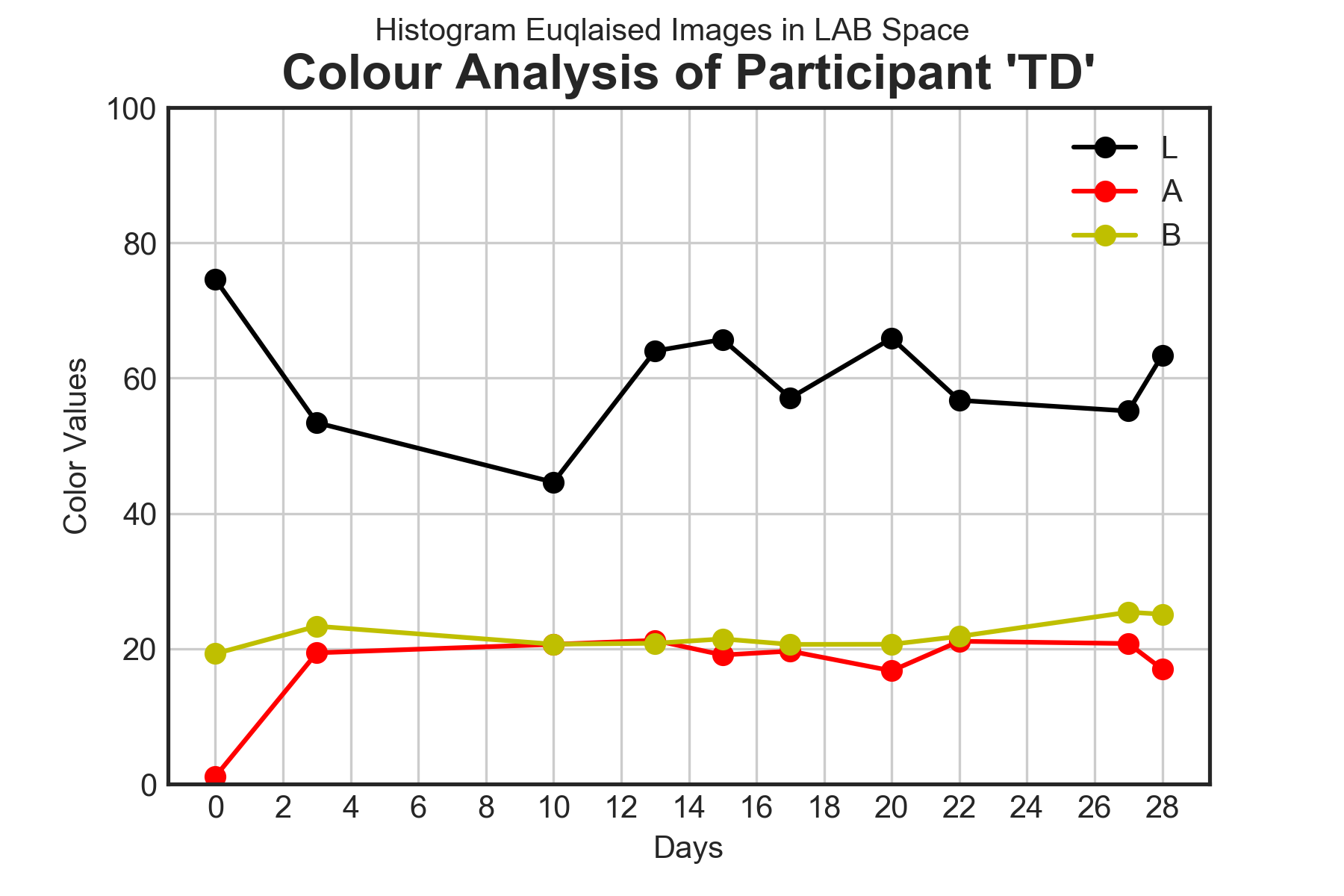}
     \vspace{0.2cm}     
    \includegraphics[width=0.33\textwidth]{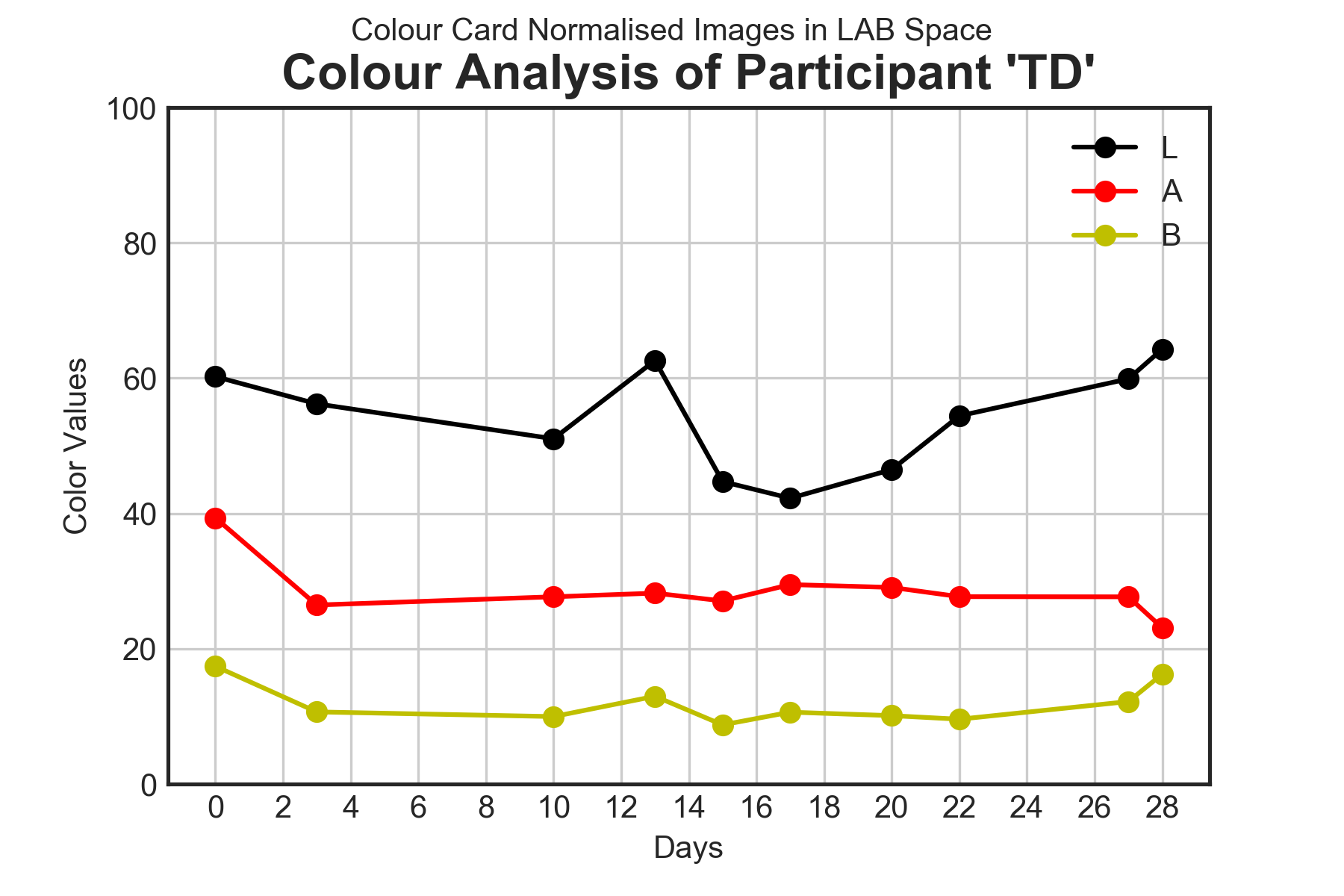}
     \vspace{0.2cm}     
    \caption{Skin Colour results for  volunteers MG, XA, HE and TD during the trial period using three forms of image normalisation}
    \label{fig:colourChange1}
\end{figure*}

\begin{figure*}[htb]
    \centering
     \includegraphics[width=0.33\textwidth]{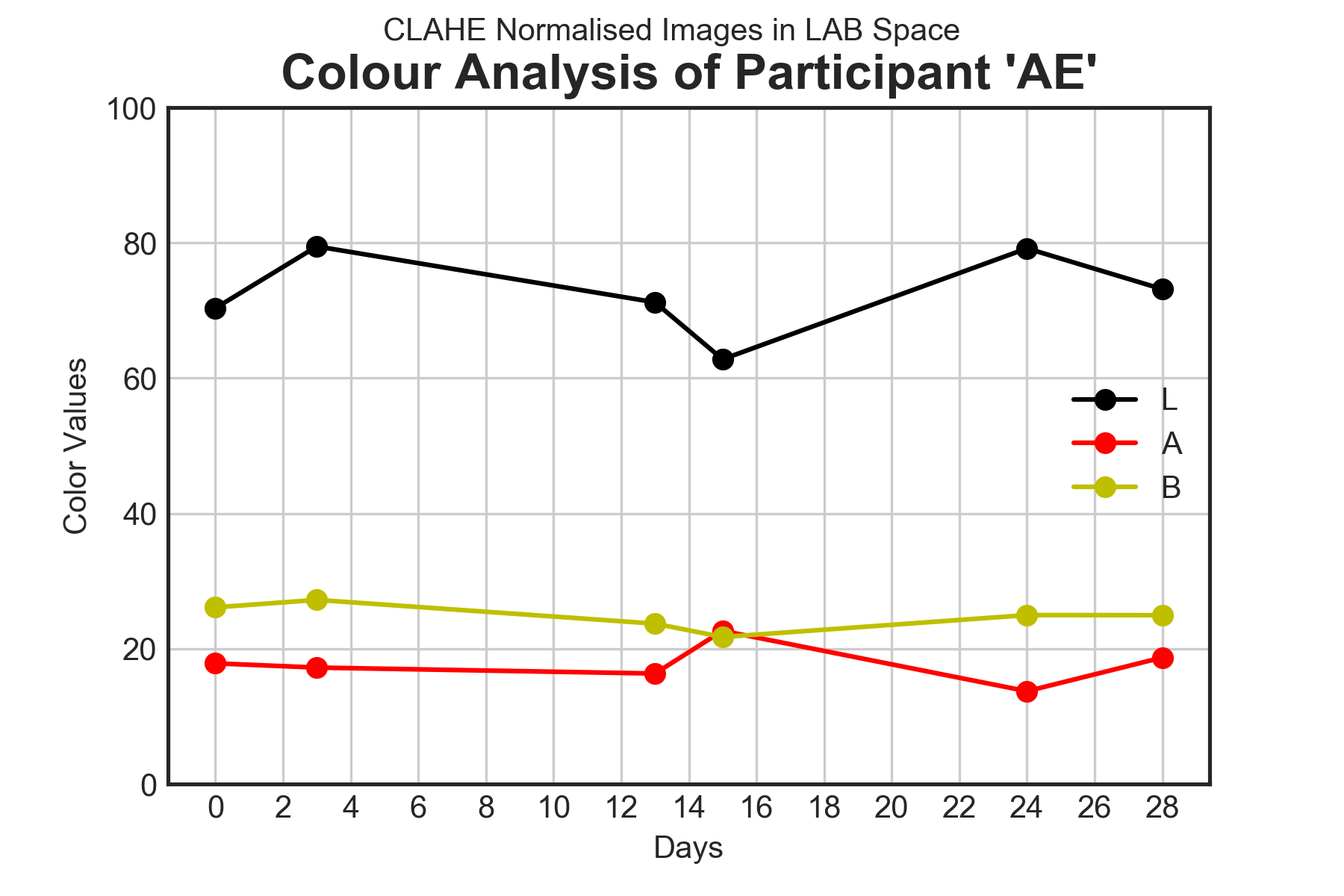}
    \vspace{0.2cm}
     \includegraphics[width=0.33\textwidth]{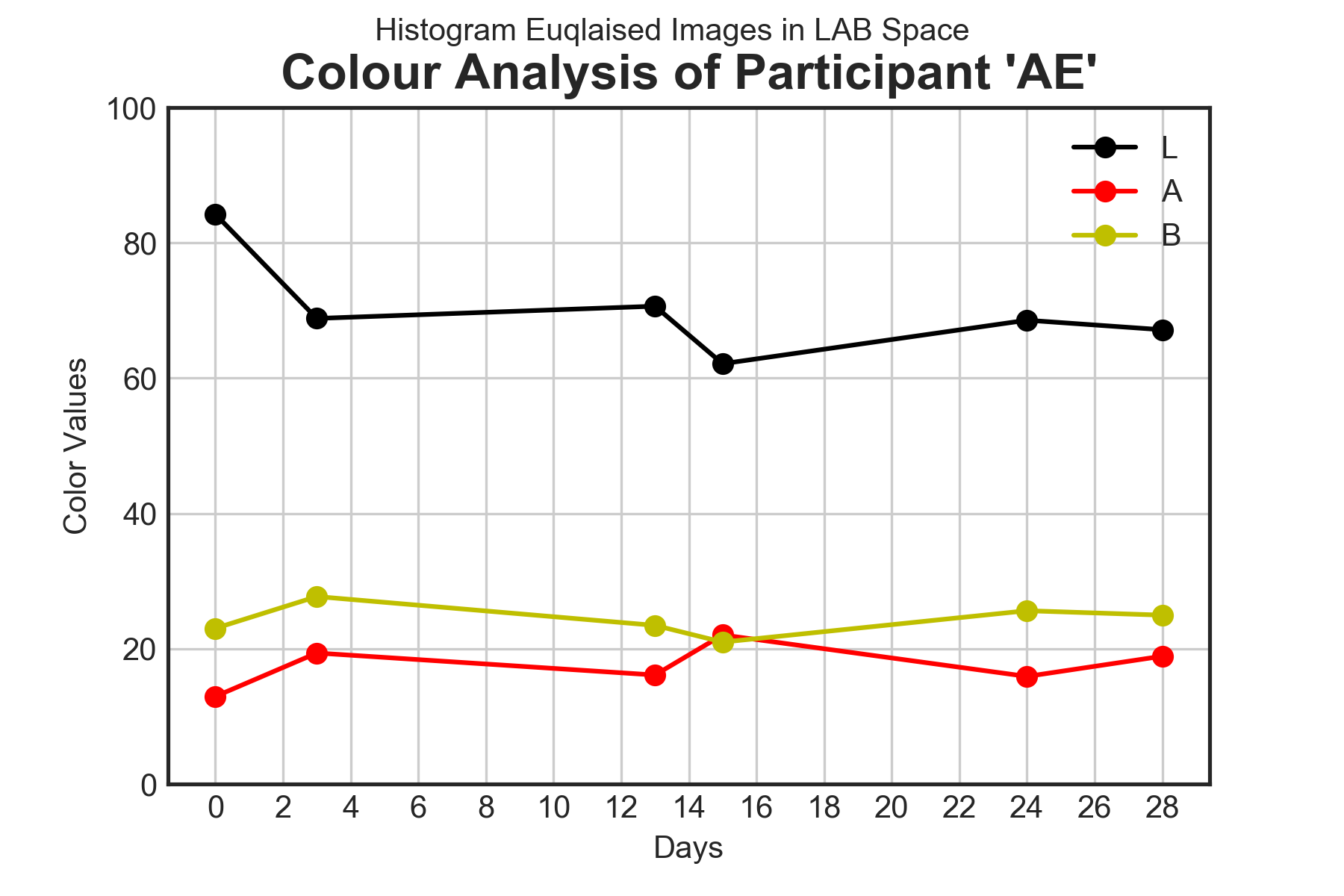}
    \vspace{0.2cm}
     \includegraphics[width=0.33\textwidth]{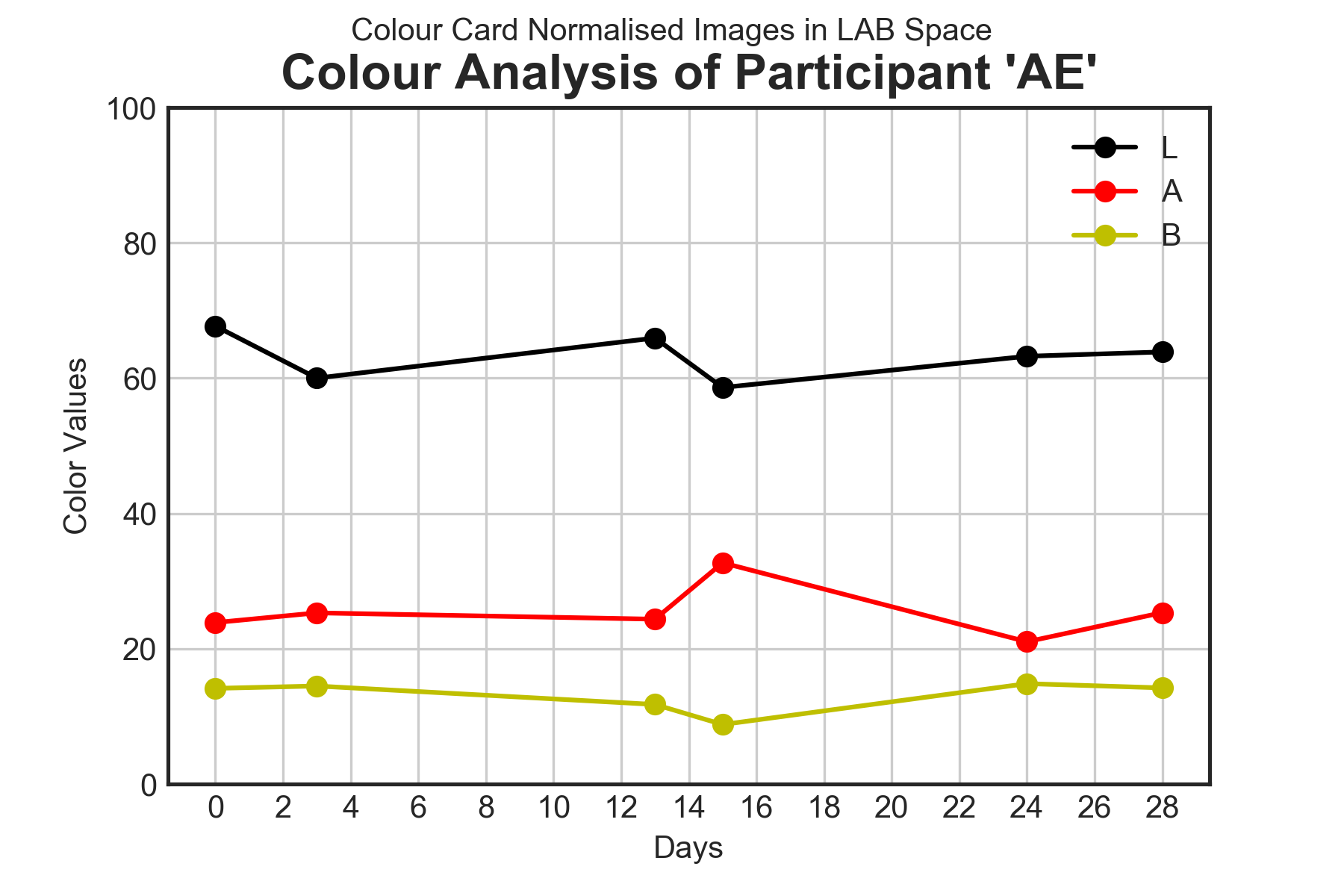}
    \vspace{0.2cm}
        \includegraphics[width=0.33\textwidth]{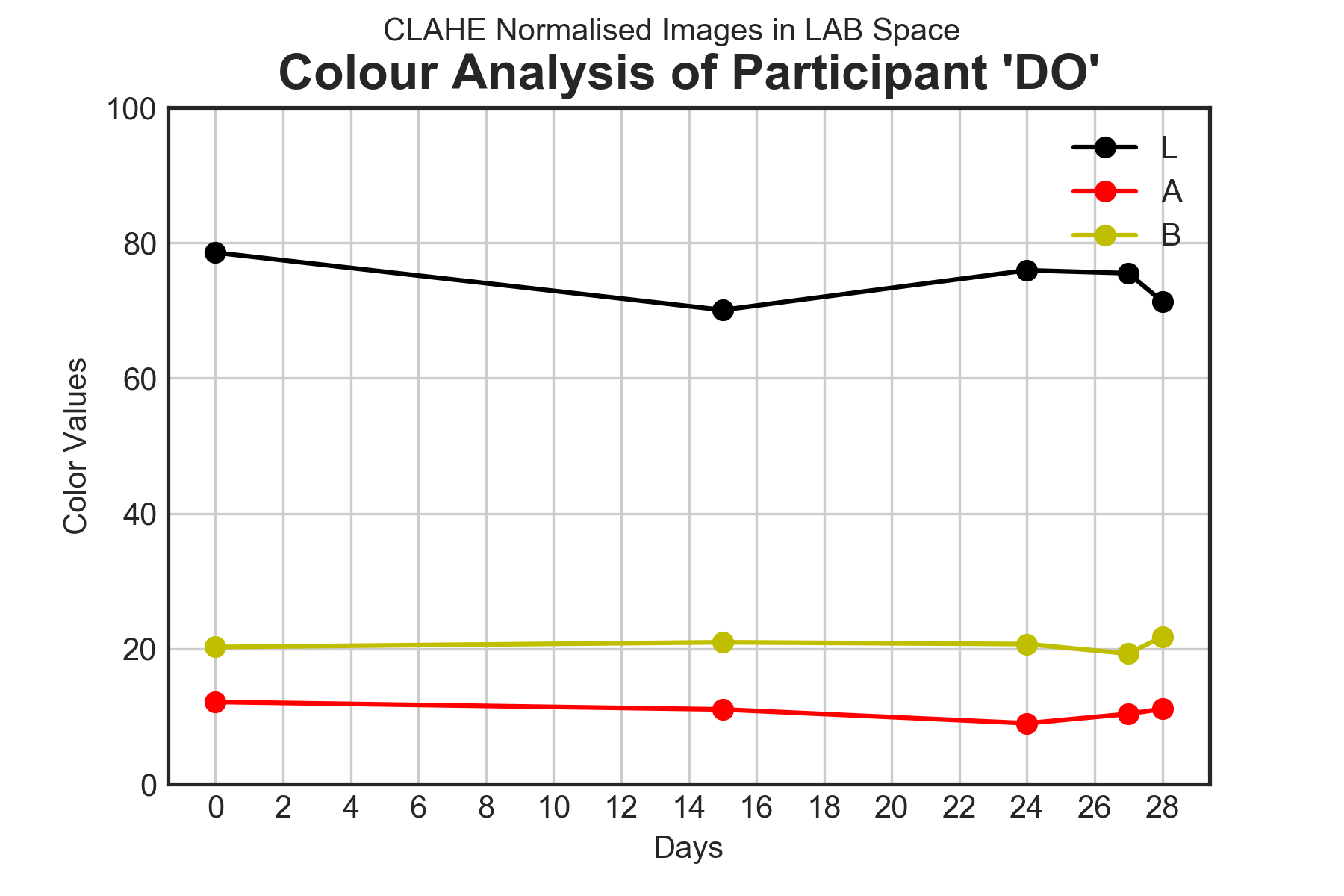}
    \vspace{0.2cm}
    \includegraphics[width=0.33\textwidth]{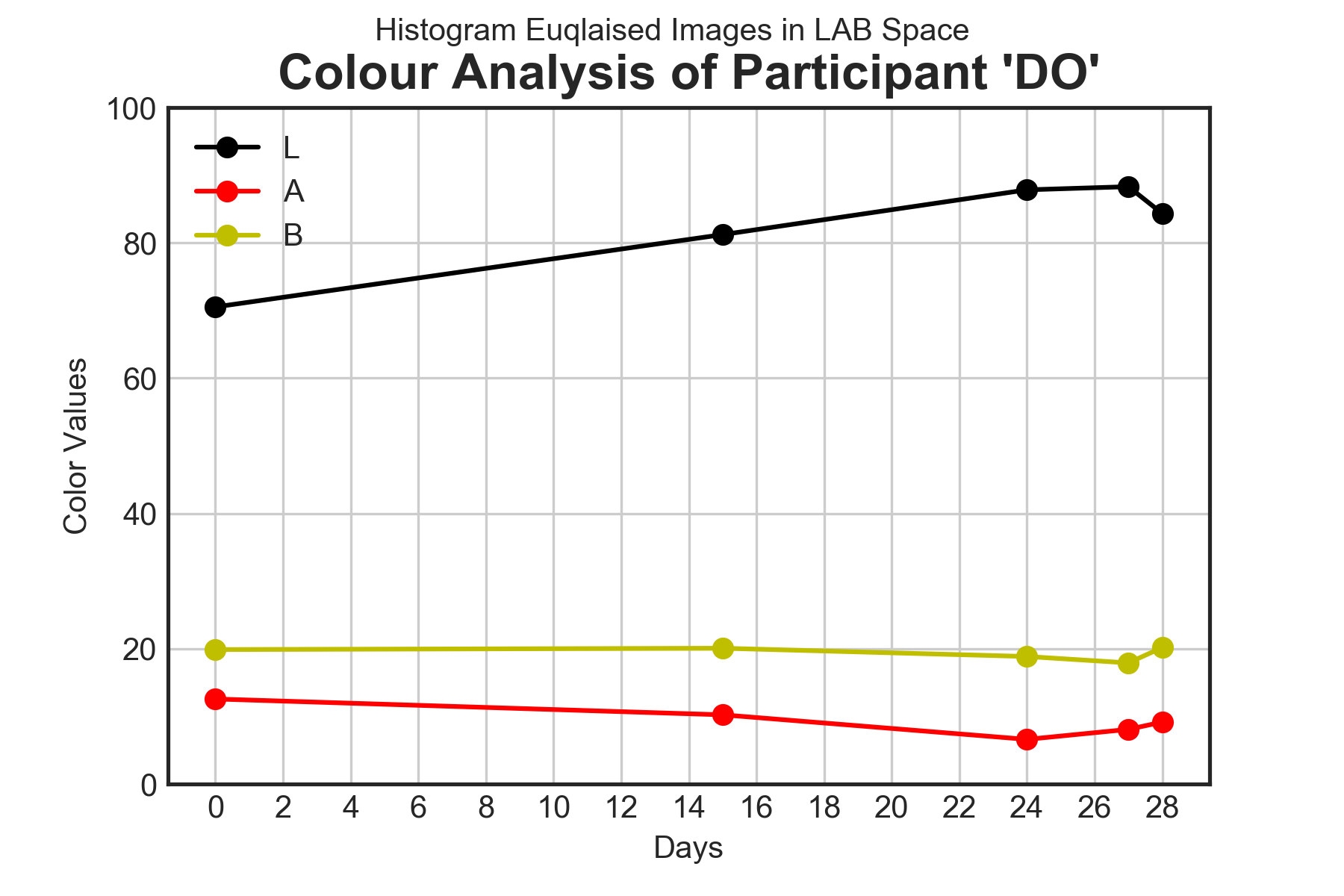}
    \vspace{0.2cm}
    \includegraphics[width=0.33\textwidth]{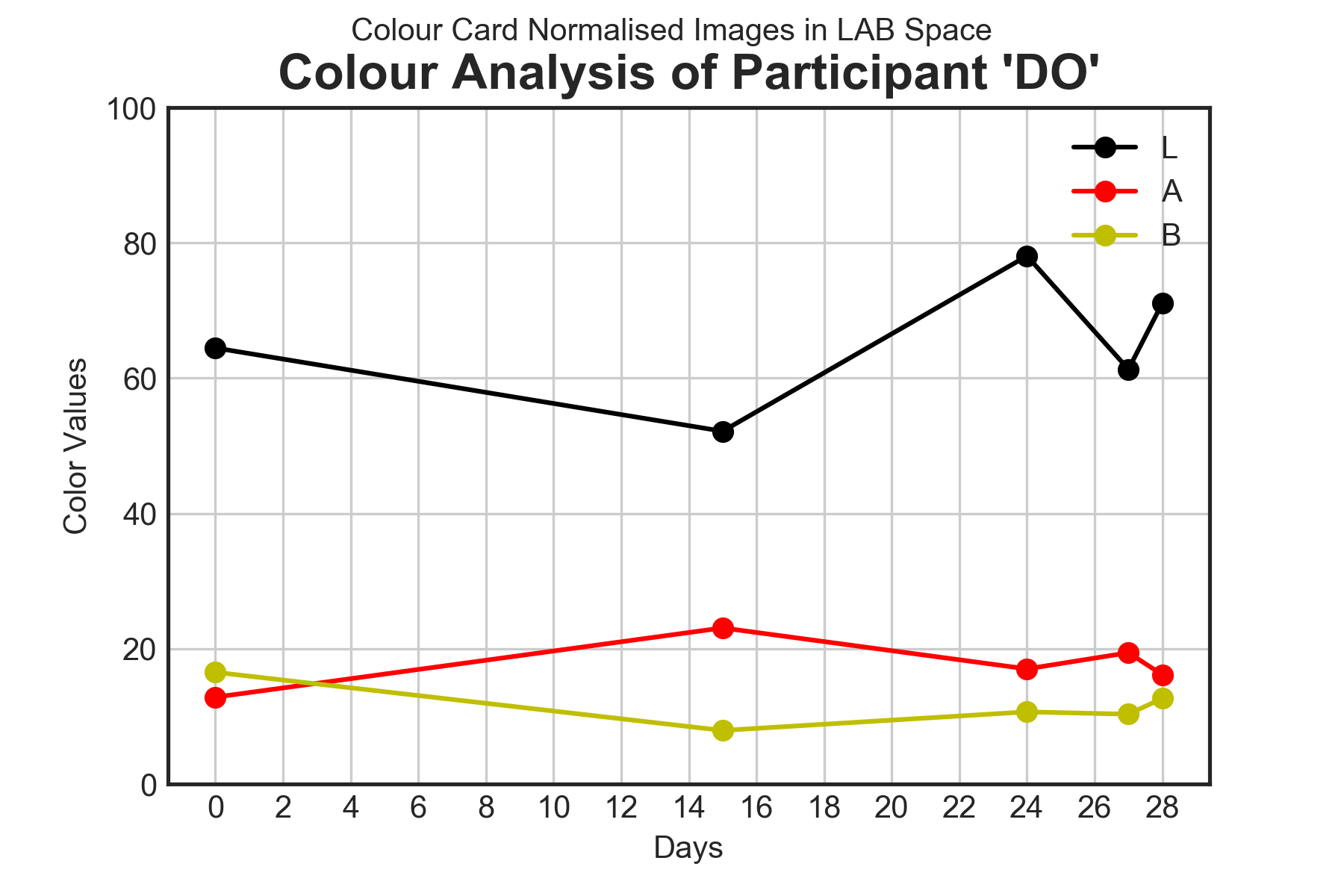}
    \vspace{0.2cm}    
    \includegraphics[width=0.33\textwidth]{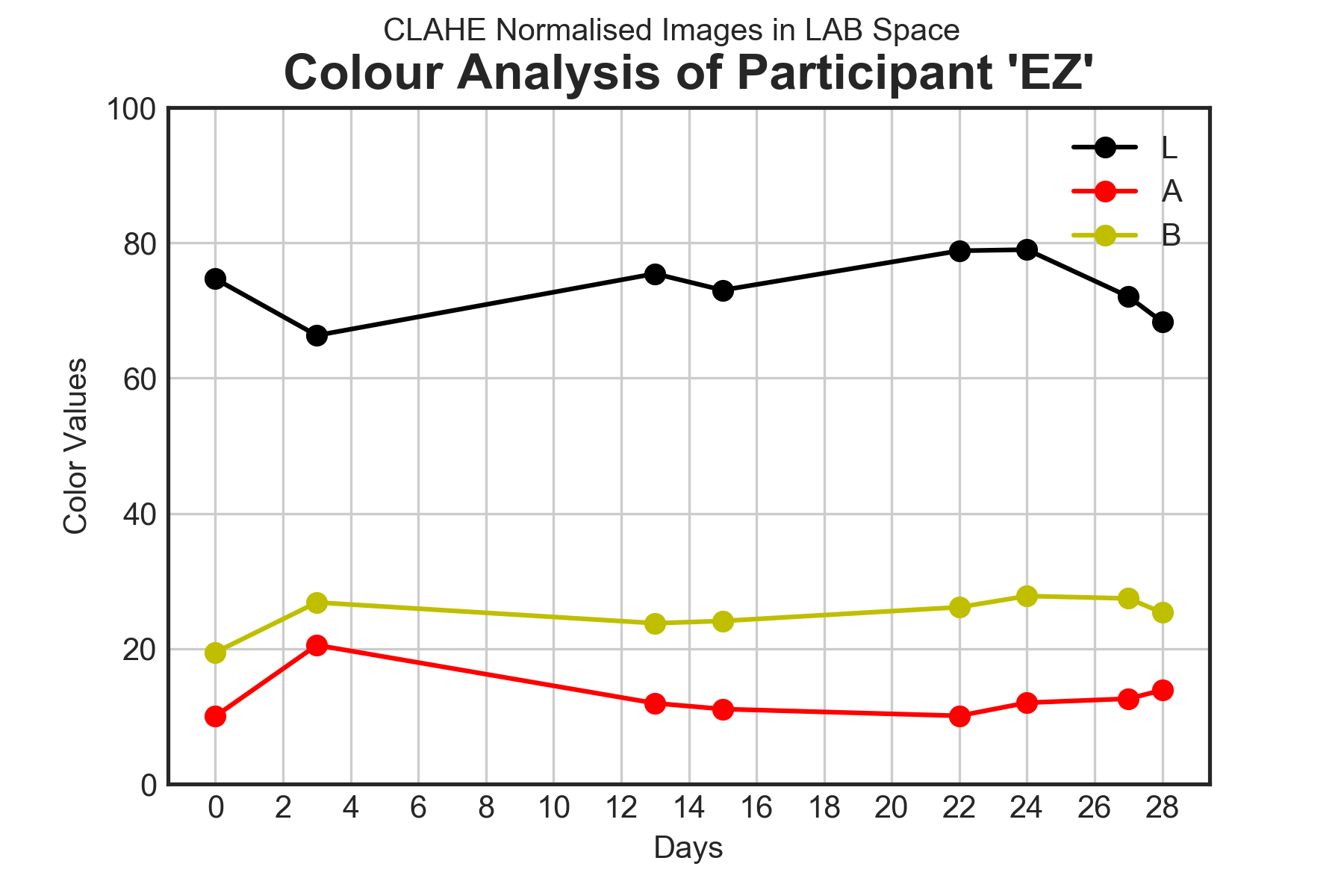}
    \vspace{0.2cm}
    \includegraphics[width=0.33\textwidth]{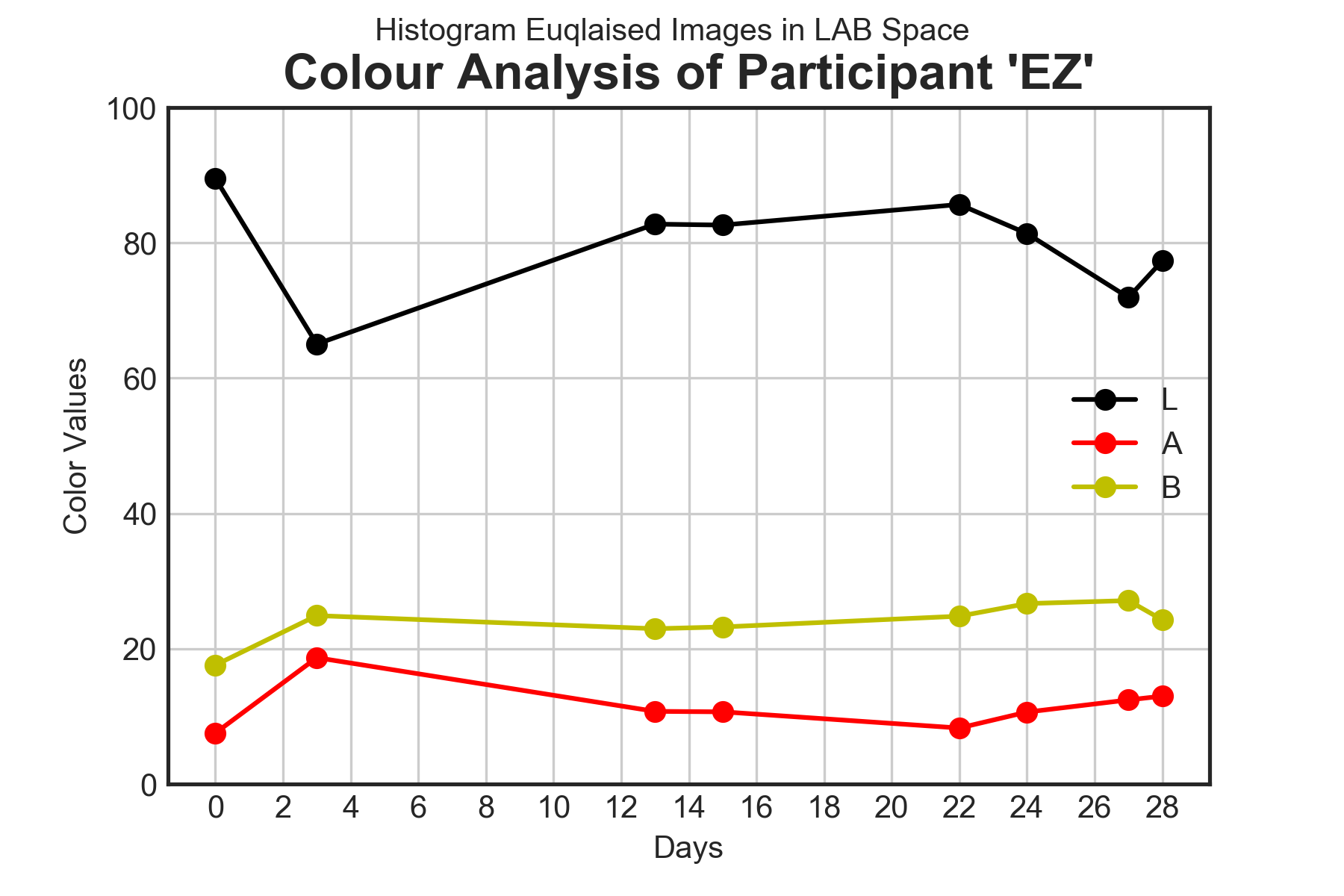}
    \vspace{0.2cm}
    \includegraphics[width=0.33\textwidth]{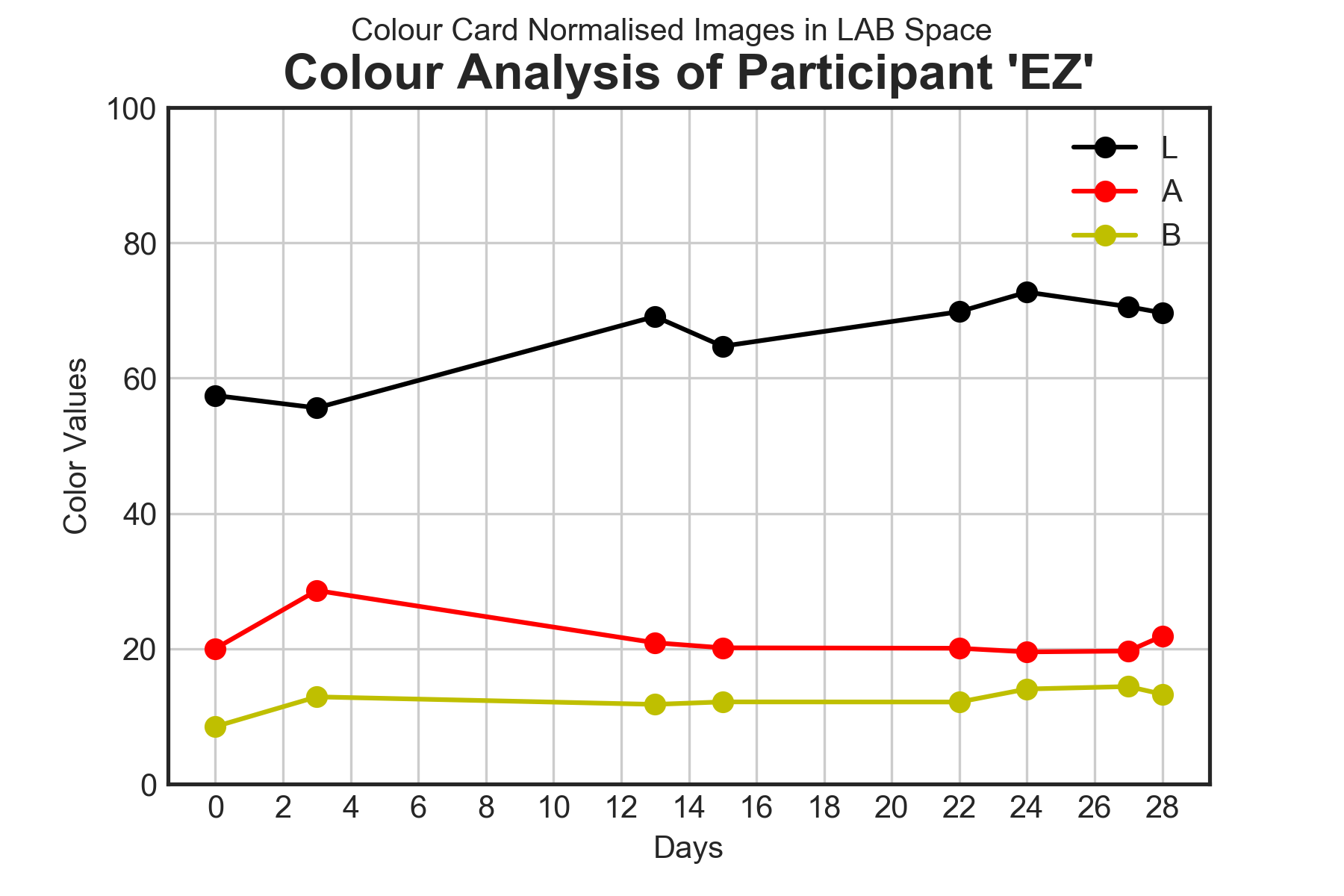}
    \vspace{0.2cm}
    \includegraphics[width=0.33\textwidth]{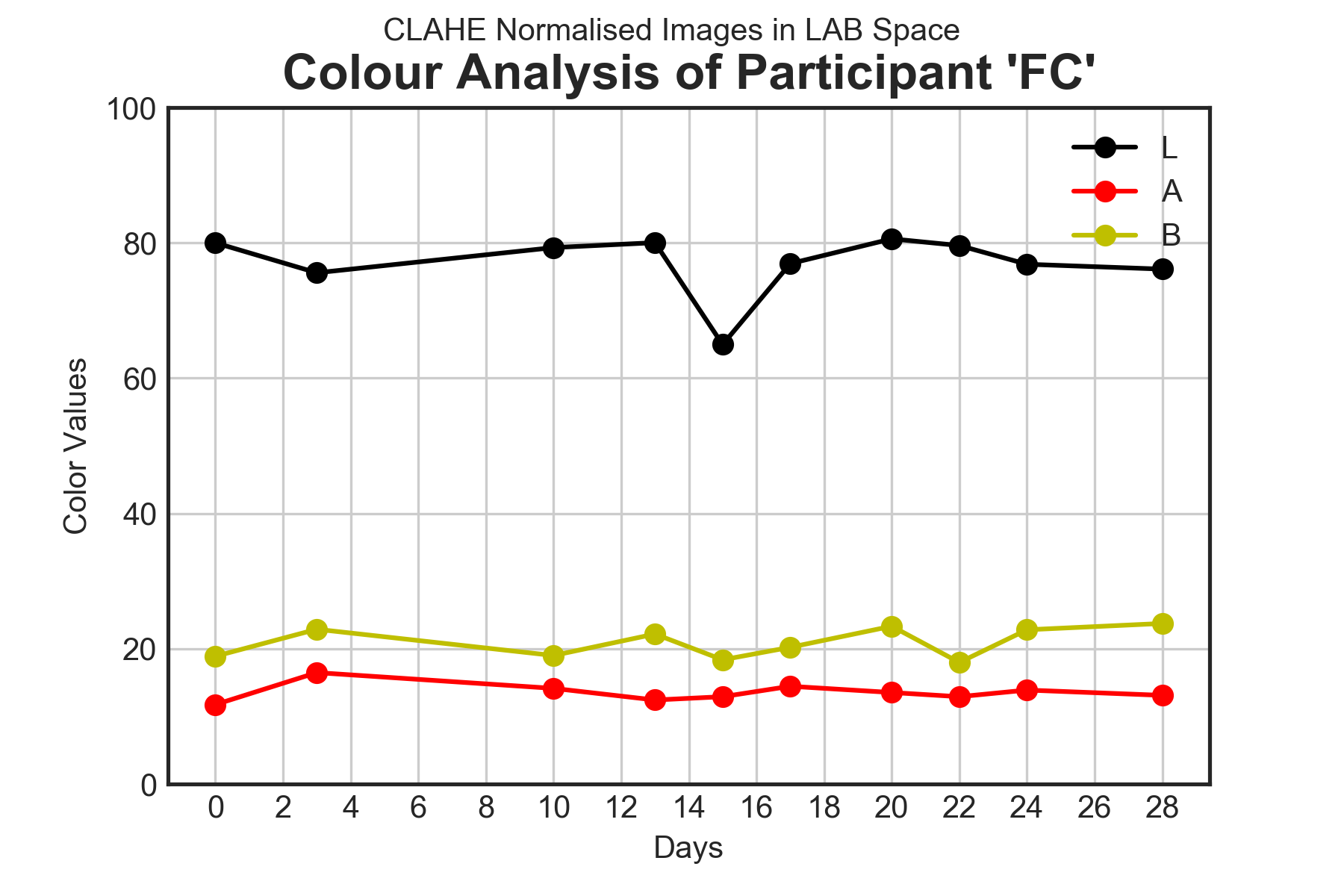}
    \vspace{0.2cm}
    \includegraphics[width=0.33\textwidth]{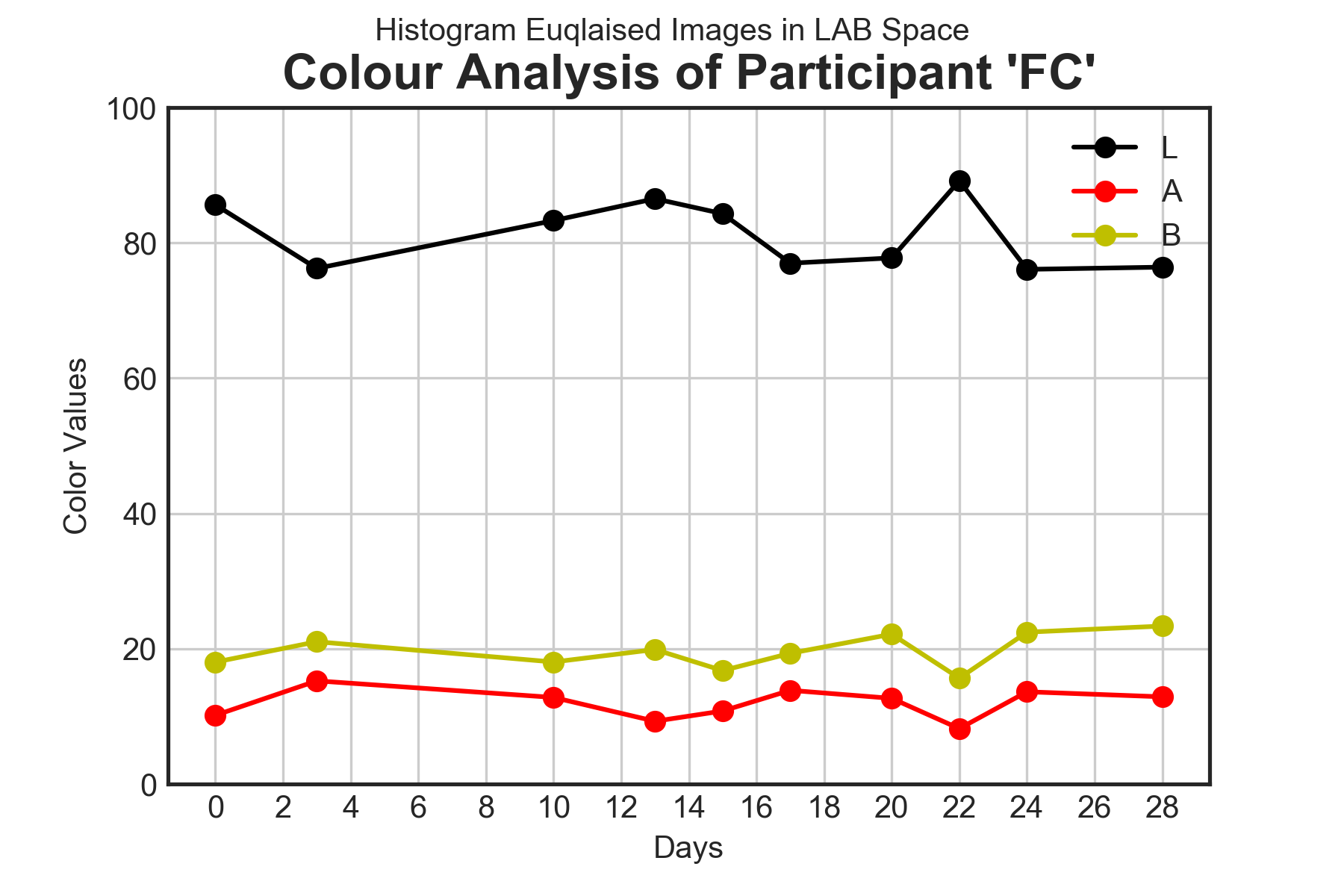}
    \vspace{0.2cm}
    \includegraphics[width=0.33\textwidth]{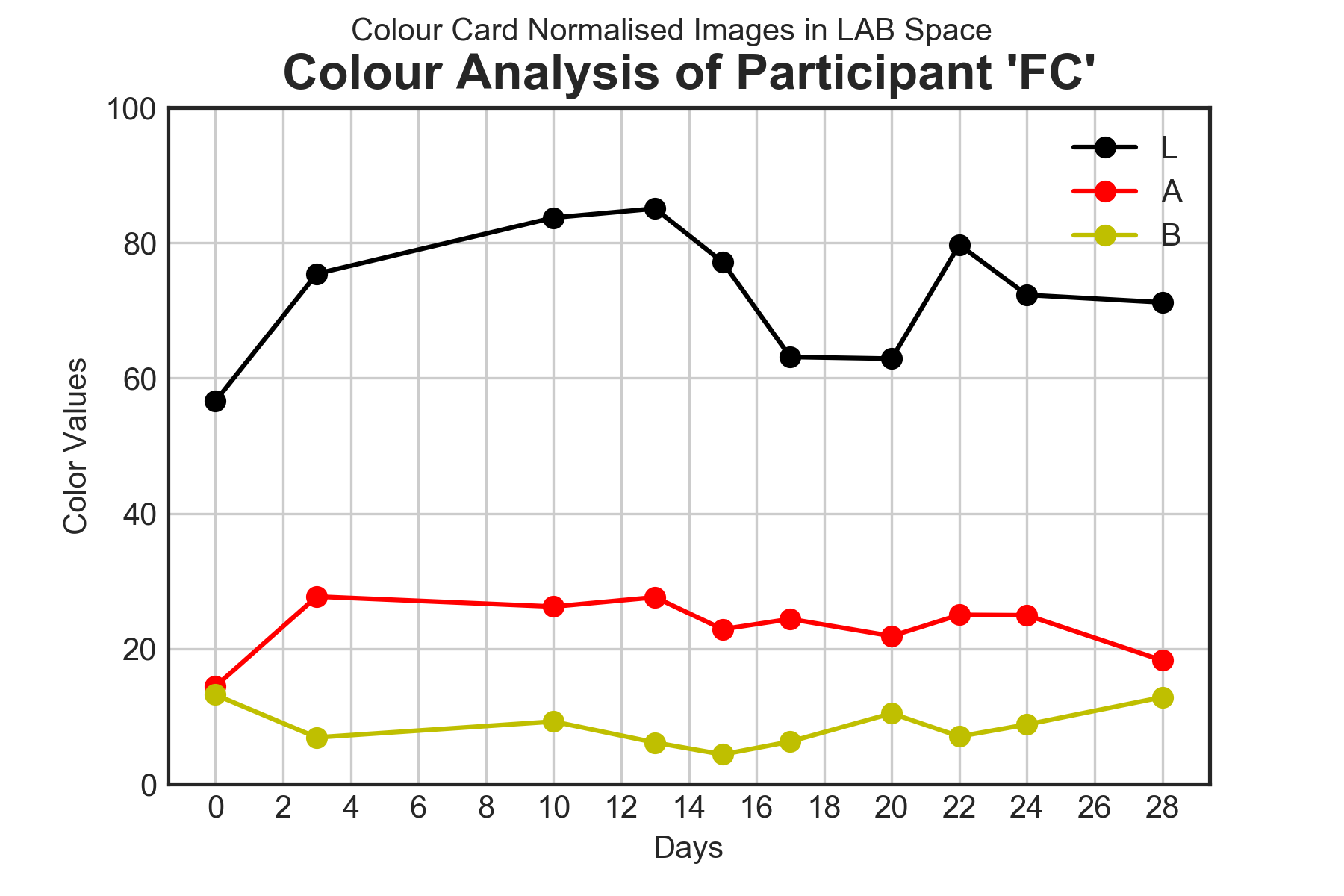}
    \caption{Skin Colour results for  volunteers AE, DO, EZ and FC during the trial period using three forms of image normalisation}
    \label{fig:colourChange2}
\end{figure*}

\begin{figure*}[htb]
    \centering
    \includegraphics[width=0.33\textwidth]{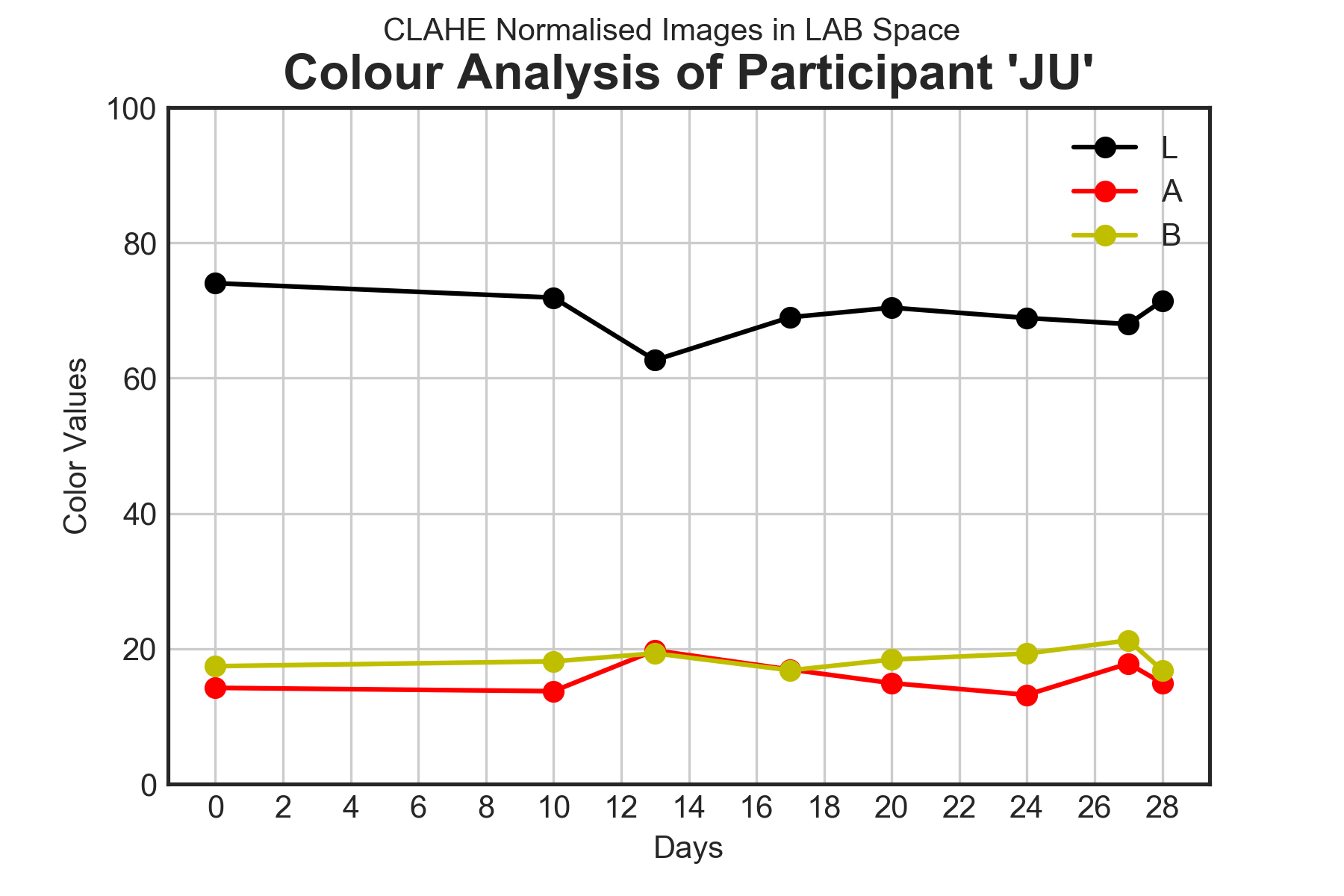}
    \vspace{0.2cm}
    \includegraphics[width=0.33\textwidth]{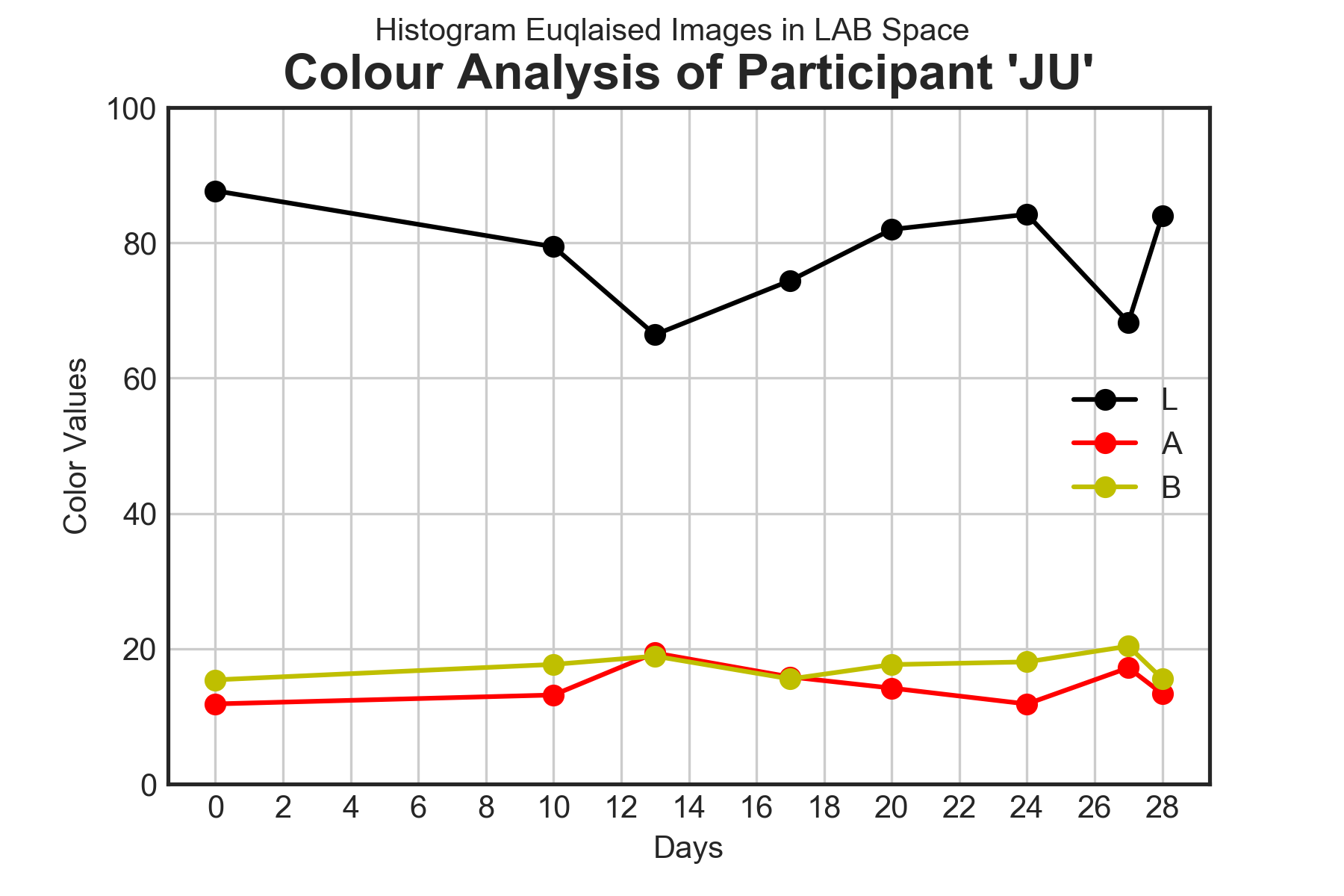}
    \vspace{0.2cm}
    \includegraphics[width=0.33\textwidth]{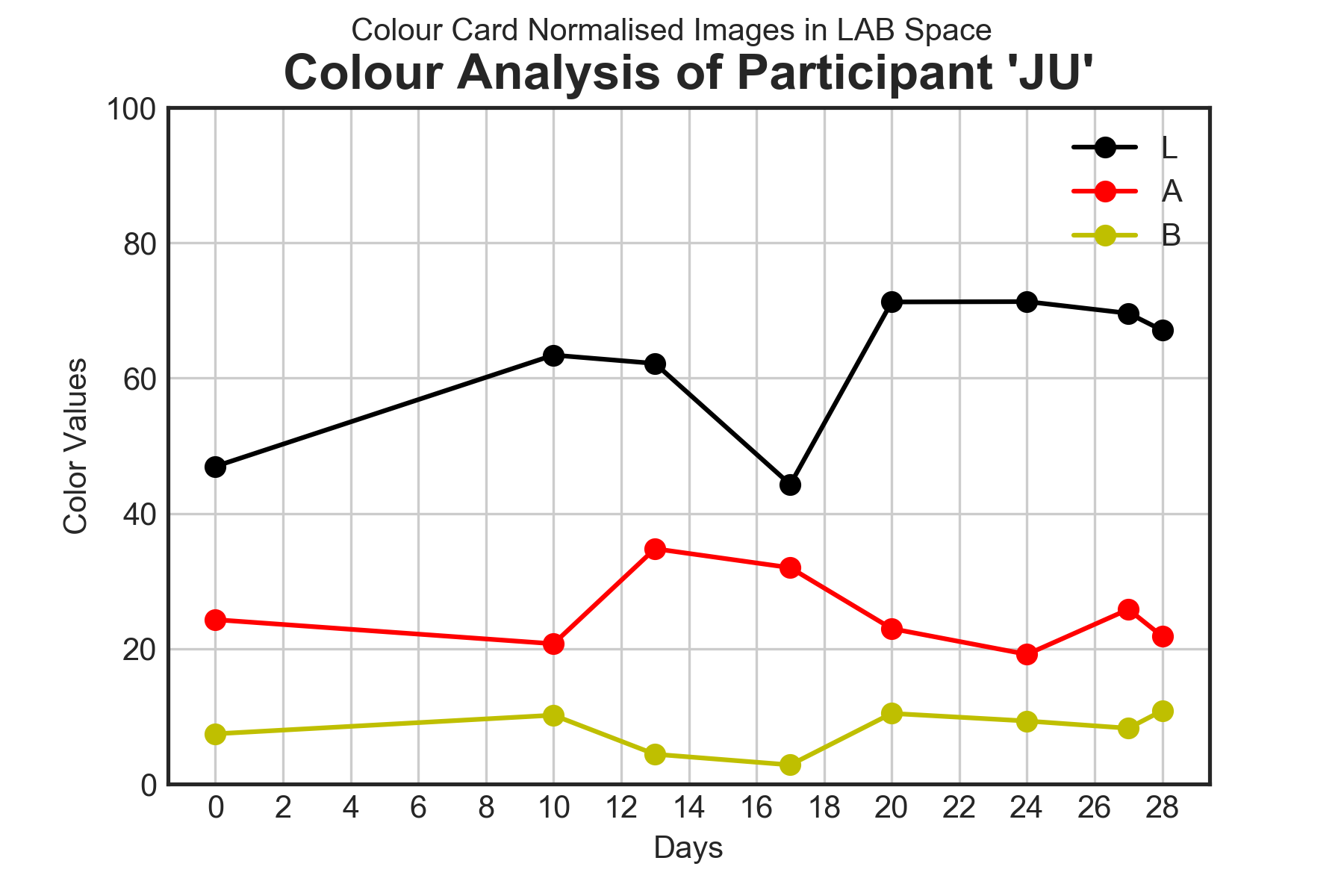}
    \vspace{0.2cm}
        \includegraphics[width=0.33\textwidth]{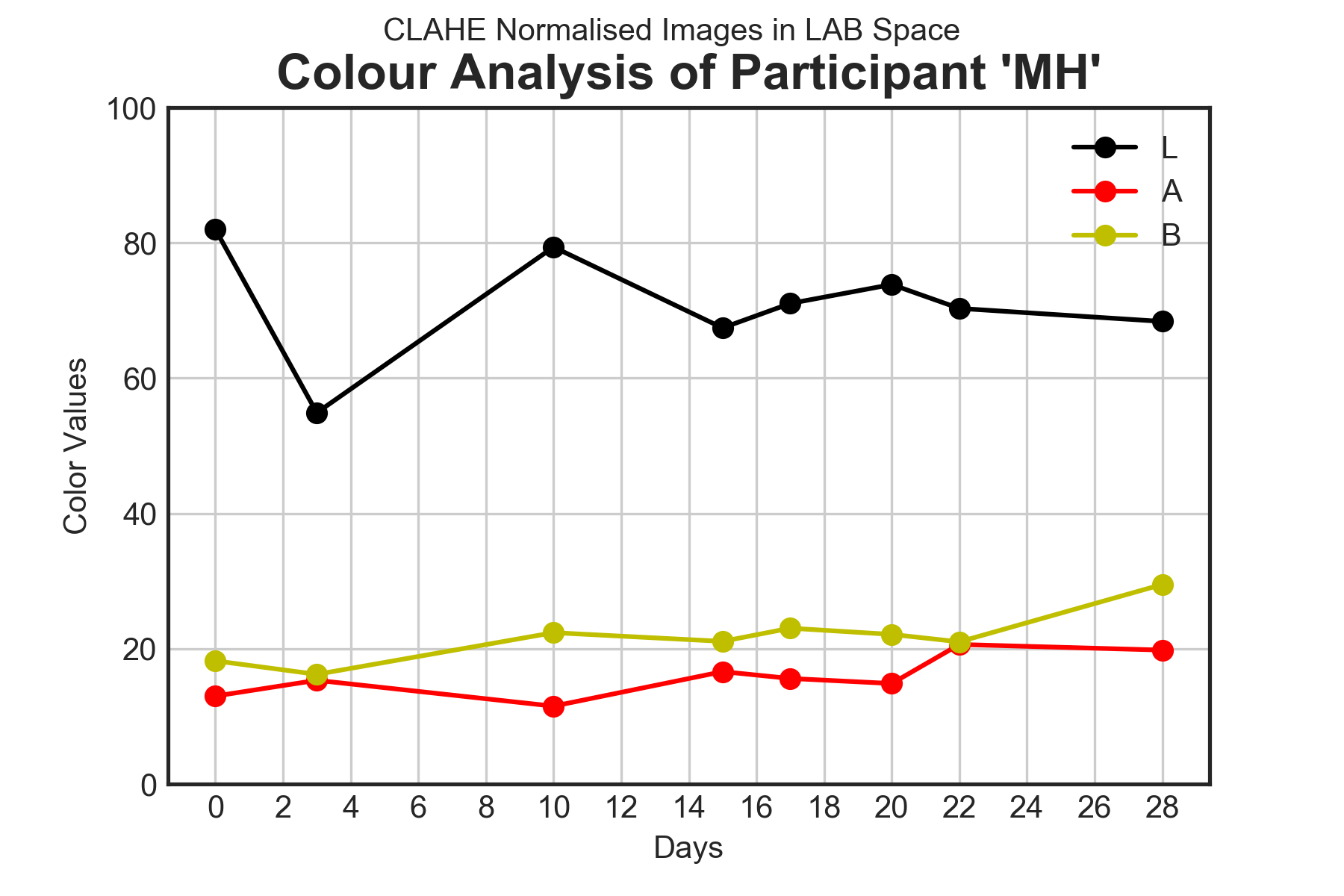}
    \vspace{0.2cm}
        \includegraphics[width=0.33\textwidth]{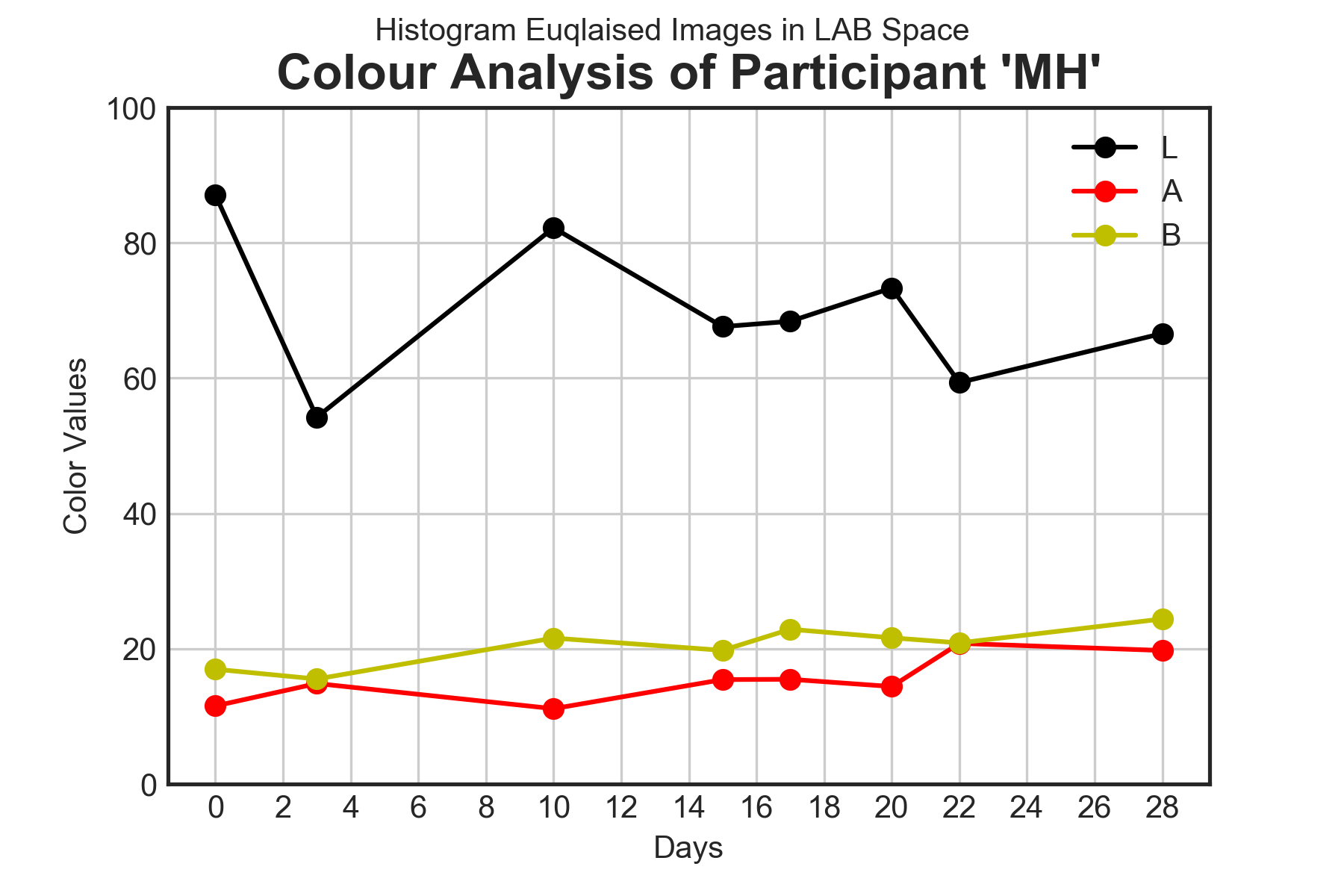}
    \vspace{0.2cm}
    \includegraphics[width=0.33\textwidth]{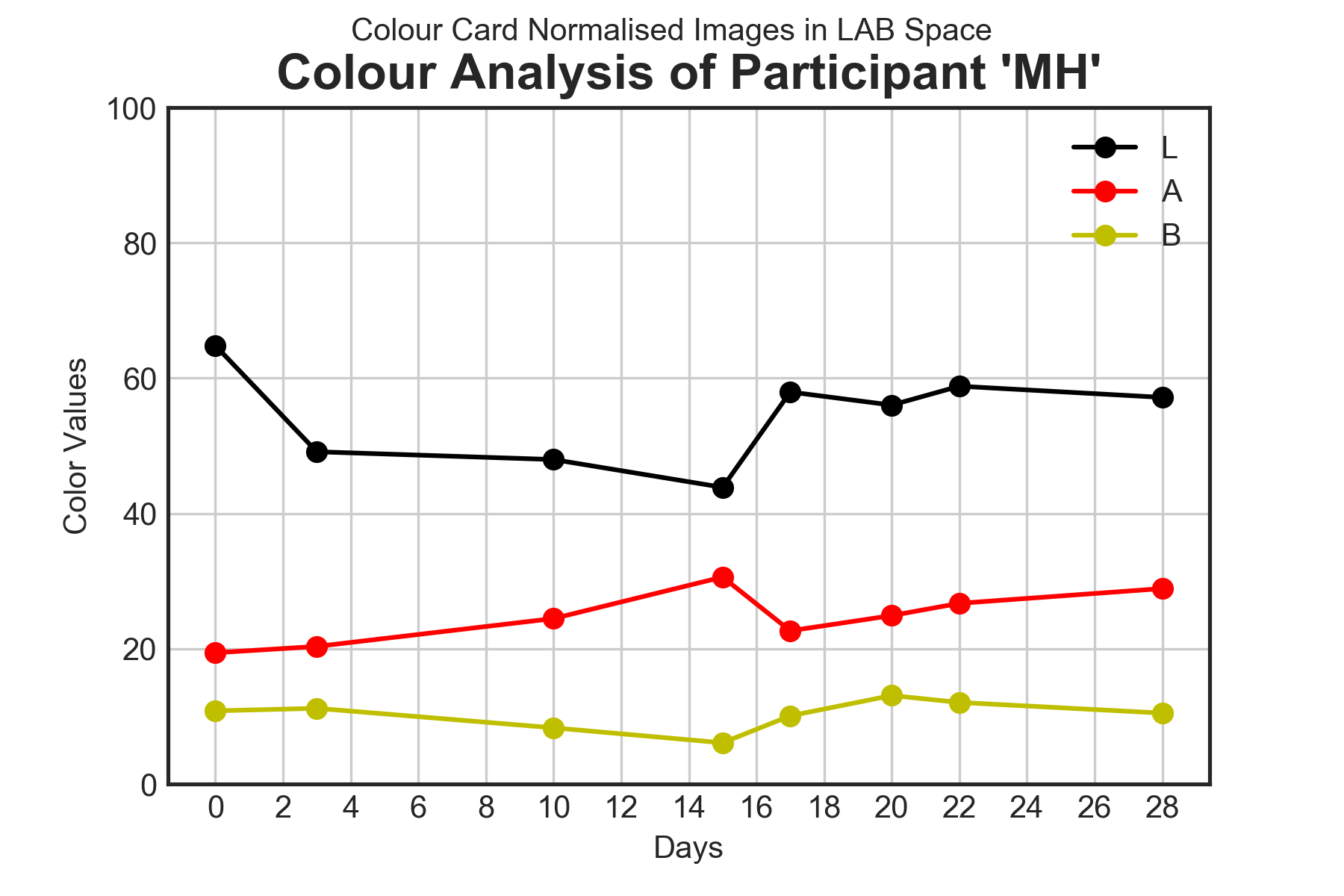}
    \vspace{0.2cm}
        \includegraphics[width=0.33\textwidth]{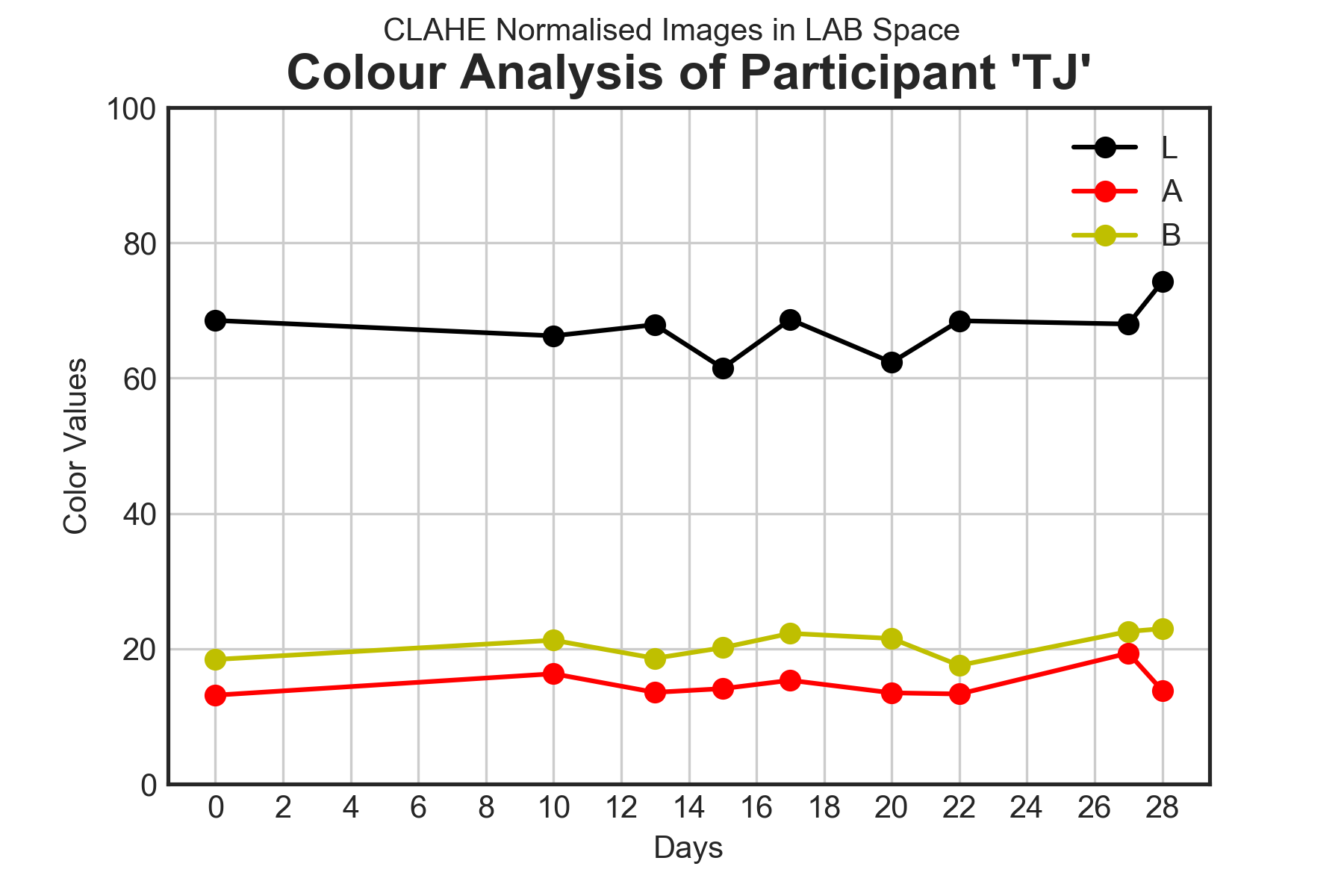}
    \vspace{0.2cm}
        \includegraphics[width=0.33\textwidth]{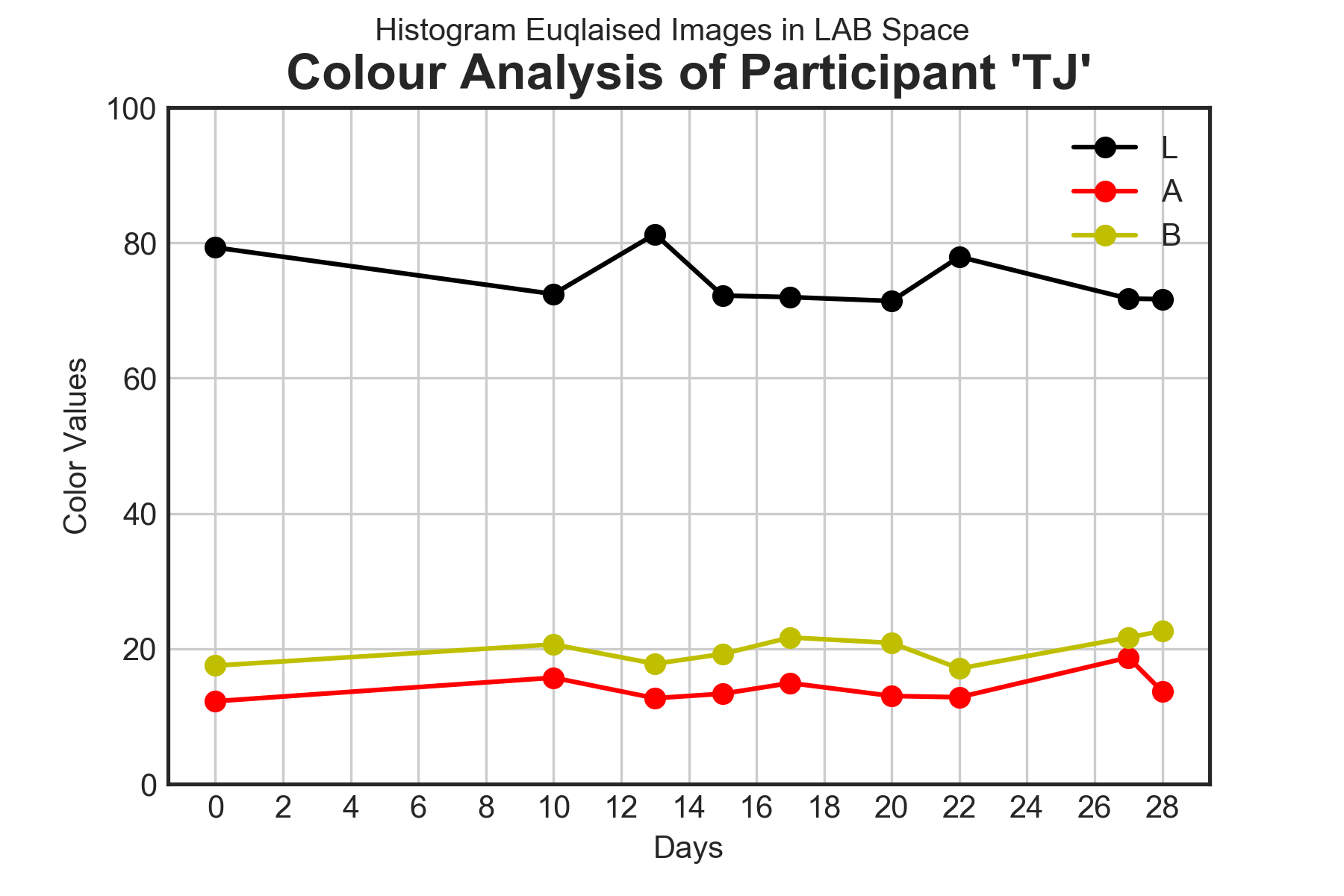}
    \vspace{0.2cm}
    \includegraphics[width=0.33\textwidth]{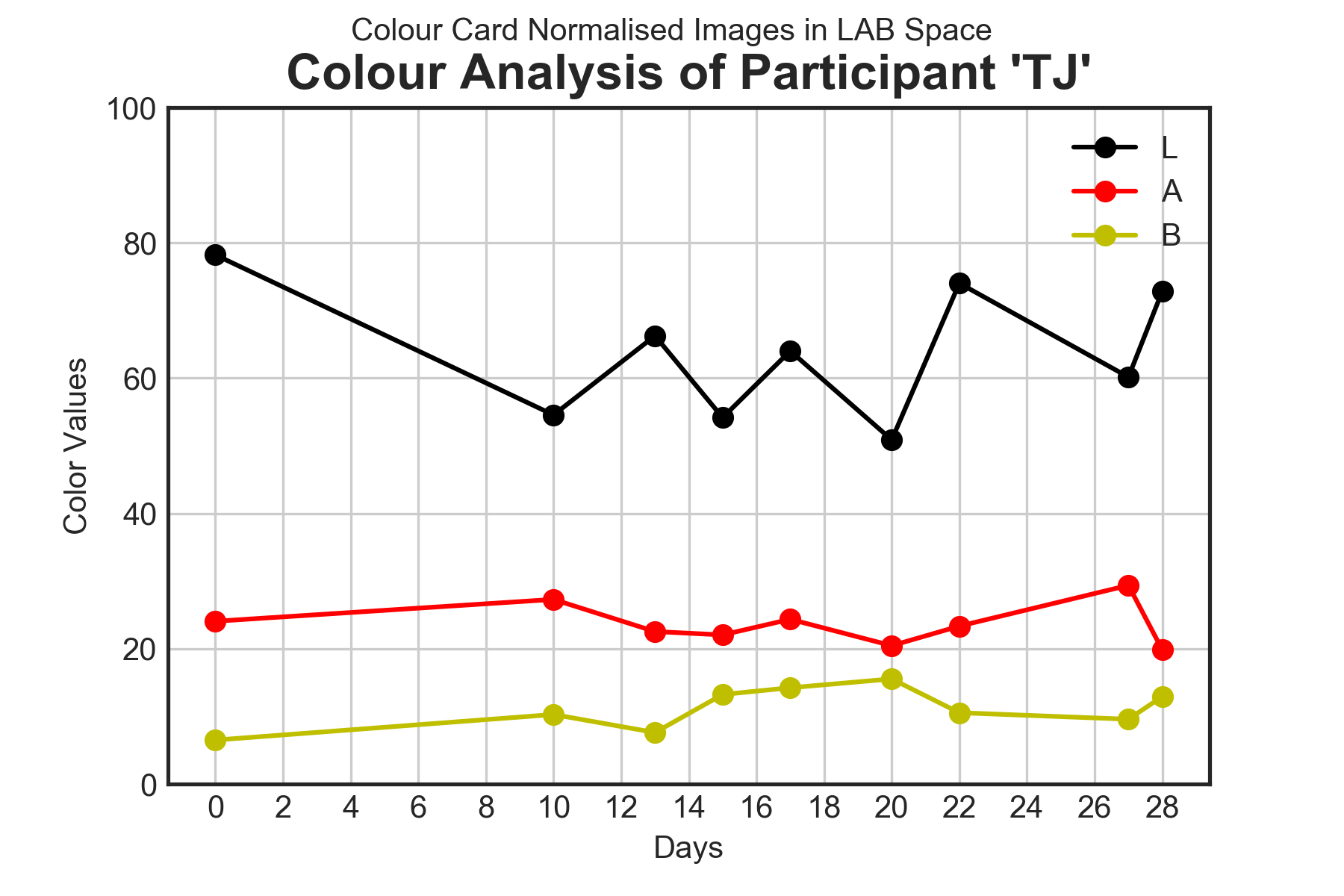}
    \vspace{0.2cm}
    \includegraphics[width=0.33\textwidth]{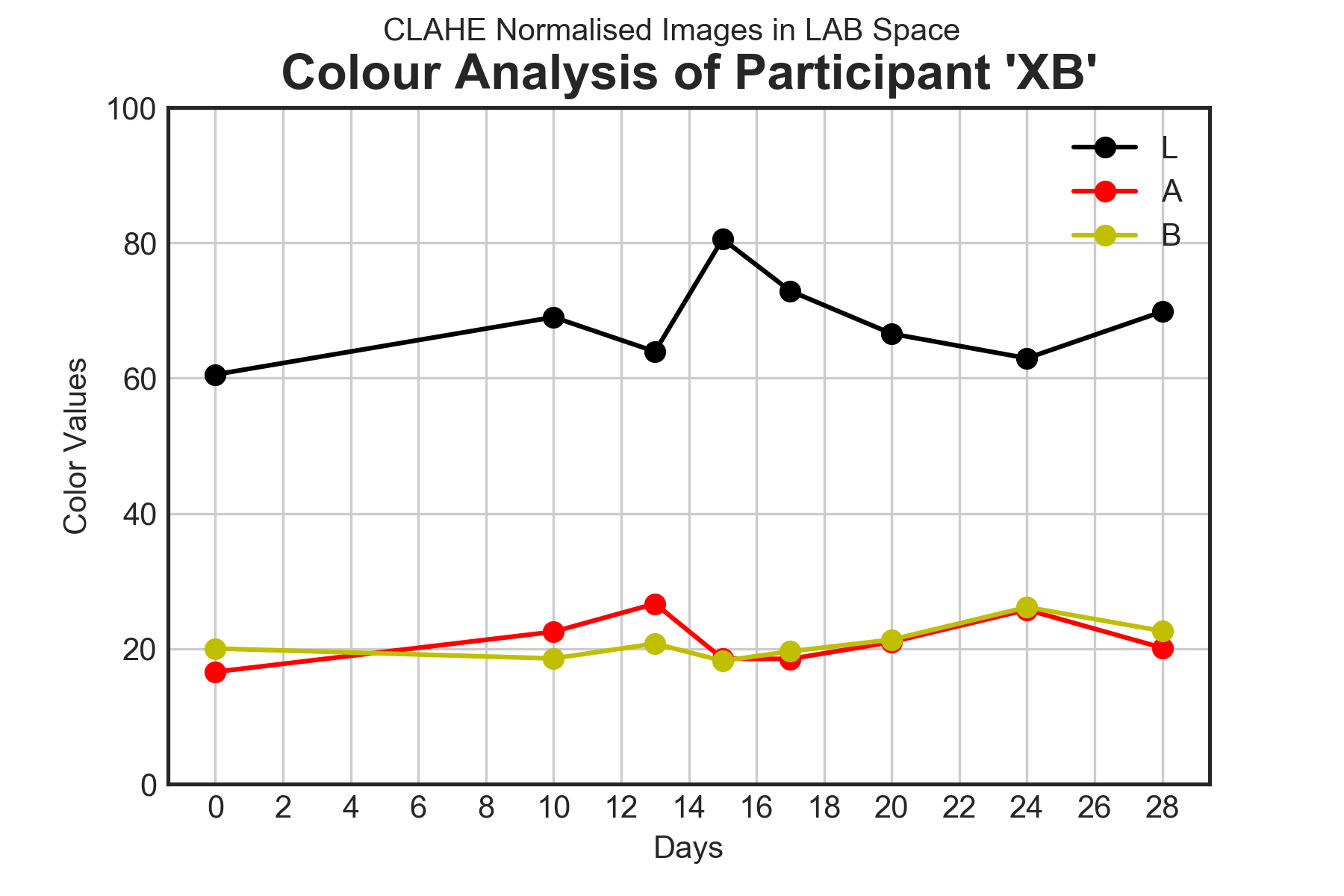}
    \vspace{0.2cm}
    \includegraphics[width=0.33\textwidth]{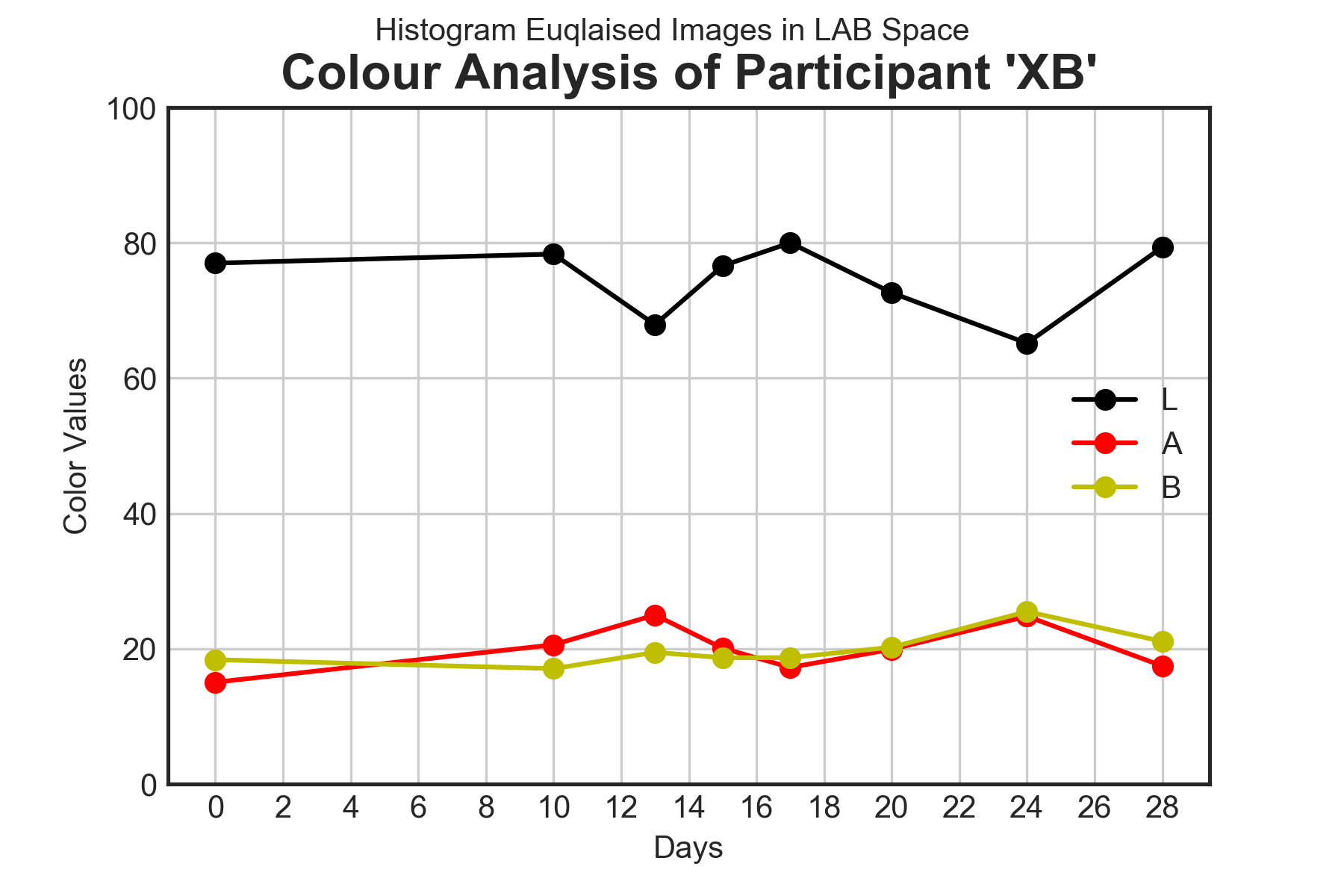}
    \vspace{0.2cm}
    \includegraphics[width=0.33\textwidth]{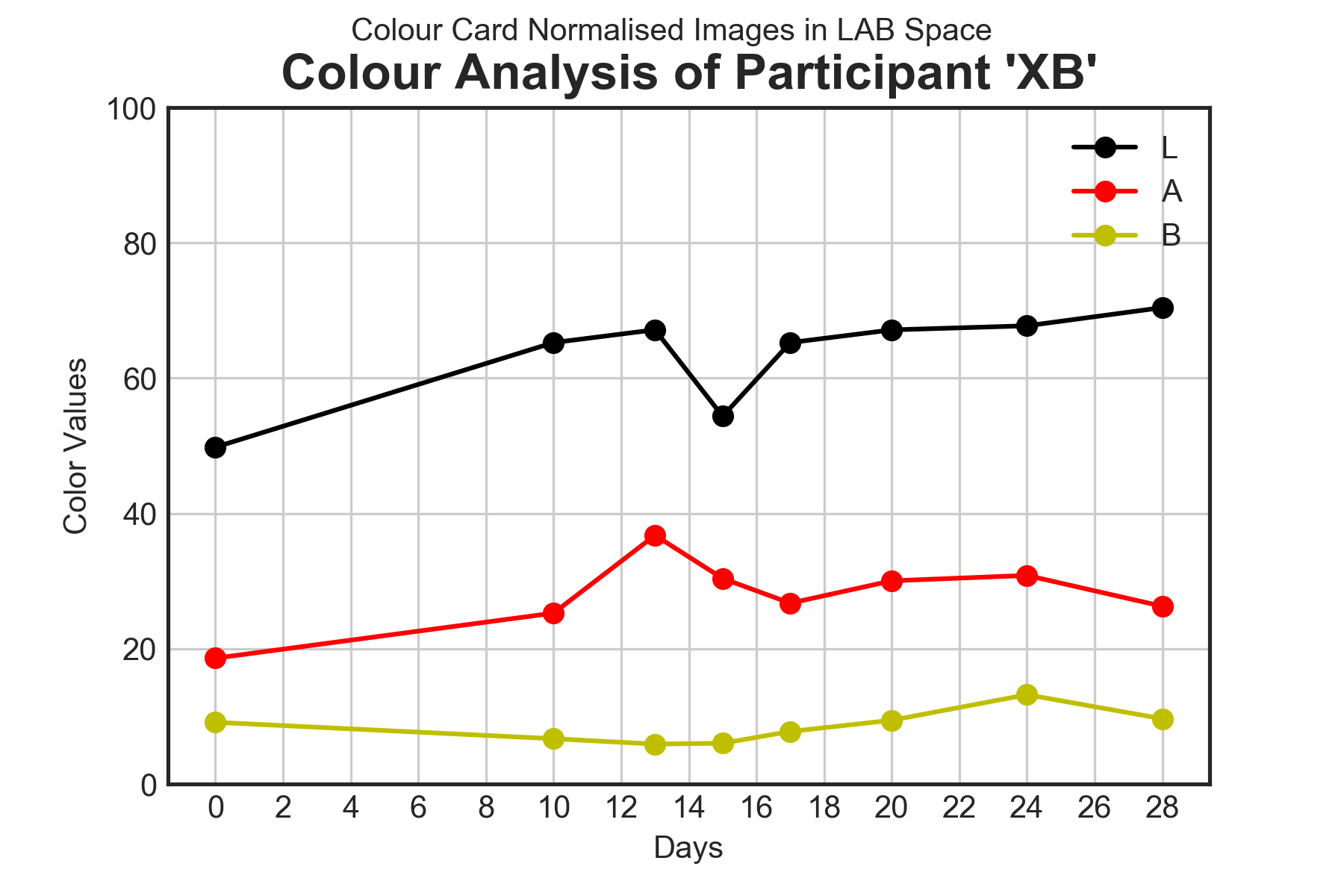}
    \caption{Skin Colour results for  volunteers JU, MH, TJ and XB during the trial period using three forms of image normalisation}
    \label{fig:colourChange3}
\end{figure*}

\noindent 
Knowing that different normalisation techniques yield different results we  visualised the colour values from each volunteer's smartphone images taken during the trial using each of the three normalisation methods.  Figures~\ref{fig:colourChange1}, \ref{fig:colourChange2} and \ref{fig:colourChange3} show the changes in each of the 12 volunteers' colour values normalised during the trial using  CLAHE, histogram-based and colour card within-image normalisation mapped to the groundtruth colour values established  from the Antera 3D camera on day 1.

The per-volunteer graphs reveal noticeable  differences across the normalisation techniques but the general shapes of the three graphs in each row in the figures are the same. Differences in the normalisation results are explained by the fact that the histogram-based technique is global across the whole image while CLAHE exploits local regions and the colour card approaches are based on known, fixed colours.
Remembering that colour values in LAB space consist of a \textbf{L} value for describing the lighting effect, \textbf{A} value for redness of the skin and \textbf{B} value for describing the yellowness we see some interesting characteristics among volunteers.  EZ, TD and DO are mostly flatlining across all colour spaces with not much change in skin colour whereas volunteers XB, XA, JU, AE and EZ each have some noticeable perturbations causing shifts in their skin colour.

\begin{figure*}[htb]
    \centering   
    \includegraphics[width=0.27\textwidth]{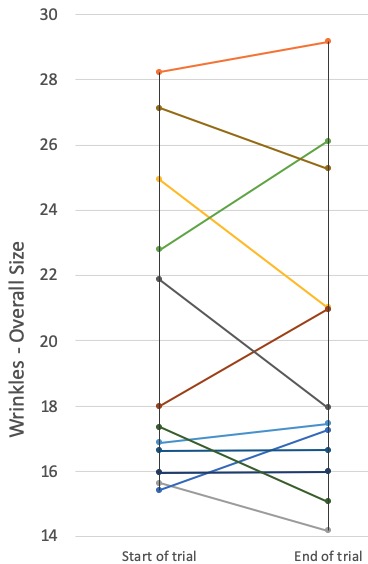}
     \hspace{0.5cm}
     \includegraphics[width=0.27\textwidth]{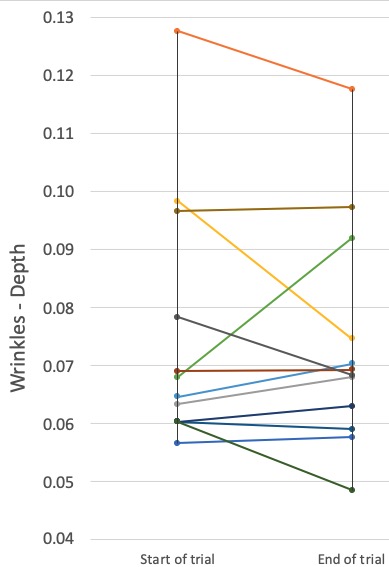}
    \hspace{0.5cm}
    \includegraphics[width=0.27\textwidth]{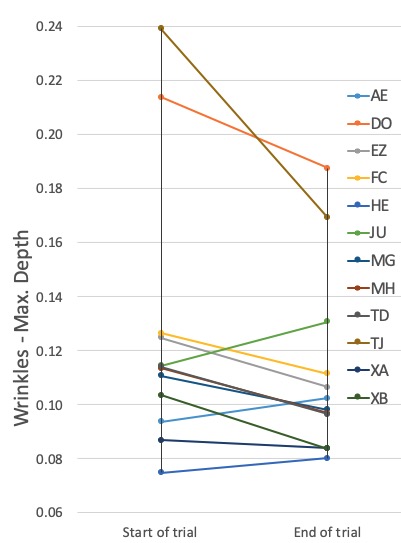}
    \caption{Skin analysis results for wrinkles for all volunteers at the start and end of the trial period using the Antera 3D camera.}
    \label{fig:lab-assessments-wrinkles}
\end{figure*}

In the second part of our experiments we focused on wrinkle analysis and 
Figure~\ref{fig:lab-assessments-wrinkles}  shows the results of the wrinkle analysis with the Antera 3D camera at the start, and at the end, of the trial period.
In Table~\ref{table:OriflameFinalResults} we saw there is an 11.2\% variation in wrinkle maximum depth between start and end of the trial with other features like wrinkle size and depth not showing significant difference.  The per-volunteer  result show that those volunteers with initially low values tended to stay low at the end of the trial (for example EZ, XA, HE, AE) while those with initially  higher values tended to reduce wrinkles (for example (FC, TJ and TD).  This is in line with what would happen physiolocally: people having a low severity of wrinkles would hardly see an improvement of their wrinkles, while people having deeper wrinkles would be far more likely to see improvements, following the usage of cosmetic products.

\begin{figure*}[htb]
    \centering
    \includegraphics[width=0.9\textwidth]{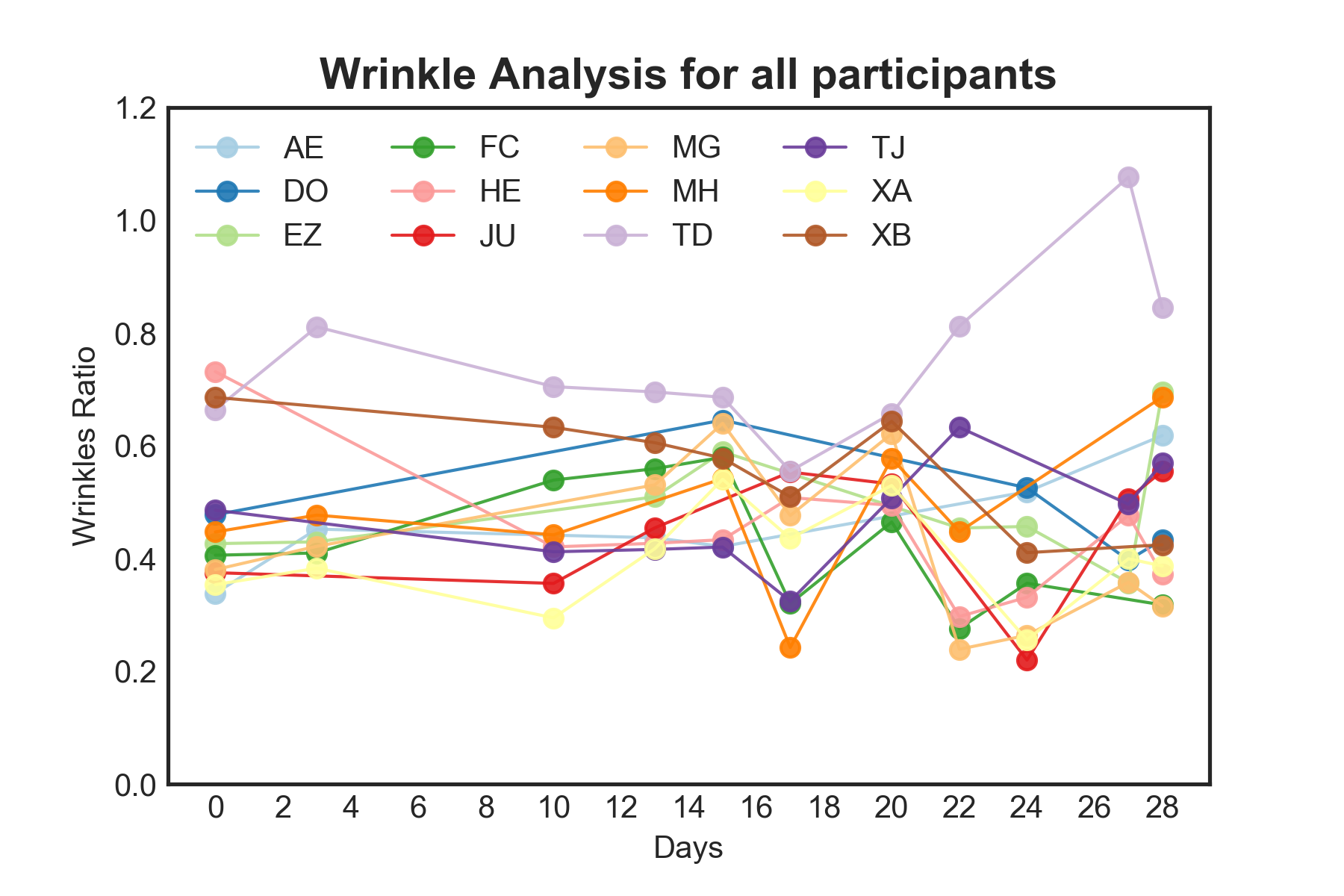}
    \caption{Wrinkle ratio results for all volunteers during the trial period}
    \label{fig:WrinkleResultGraph}
\end{figure*}

Using the within-trial smartphone images we identified the wrinkle area and computed its gradient magnitude and Figure~\ref{fig:WrinkleResultGraph} shows the changes in wrinkle ratio per volunteer observed during the trial period.  
If the appearance of wrinkles in the image is greater, then more edges will have been detected by the gradient operator. In such a case, the value from the Sobel combined is large, resulting in a higher value for Wrinkle ratio. 
What this analysis shows is that almost all volunteers have some perturbations in their wrinkle anaysis during the 4-week trial and there are no``flatlines" in the graph in Fugure~\ref{fig:WrinkleResultGraph}.  Between days 18 and 18 (which was May 17th, a Friday) something happened to impact everybody's assessment. As with the analysis of skin colour, follow-up interviews with individual volunteers would be needed to explore the causes for these changes but once again we see that the analysis of smartphone images taken during the trial gives insights into within-trial changes of volunteers' skin.


\par
\section{Conclusions}
\label{sec:Conclusion}
This research aimed to study the intermediate progress of cosmetic lab product trials using smartphone images so as to gather additional information during the trial period allowing better analysis. Working with a cohort of 12 female volunteers we focused on skin colour which required colour normalisation on the images to adjust for varying daylight conditions, and automatic analysis of wrinkles based on measuring lines in the greyscale version of the images.
We used images from an Antera 3D camera taken at start and end of the trial  and then images taken from a consumer level  smartphone taken throughout the trial period.

When we examine per-volunteer changes in both skin colour and wrinkle ratio we see that for many of the volunteers there are noticeable perturbations in both their skin colour and wrinkle ratio indicating non-linear transitions from the start to end points.
This clearly highlights the contribution of other confounding variable that can not be controlled/recorded accurately as part of regular clinical trials with fixed appointments: weather fluctuations,  food intake, mood, sleep, exercise and so on.
A face-to-face exit interview with each volunteers, or the volunteers keeping a log or a diary of events during the trial would allow  those conducting the product trial to get a  deeper understanding of skin changes not just at the two fixed timepoints of start and end, but throughout the trial.

The results obtained in this paper indicate there is potential for using within-trial facial image data gathered directly by participants using a consumer-level smartphone.  While there are limits on the ability of such devices to capture detailed high resolution, 3D facial surface textures such as described in \cite{doi:10.1111/srt.12793}, the correlation between normalised smartphone images and the Antera 3D camera for skin colour parameters are very encouraging.

Lifelogging represents a phenomenon whereby people can digitally record their own daily lives, ordinary lives, in varying amounts of detail, and for a variety of purposes \cite{gurrin2014lifelogging}. This can include using simple wearable devices like fitness trackers, step counters and sleep monitors, right up to front-facing cameras and wearables to record heart rate, respiration, stress, etc.
Including lifelog data into clinical trial analysis, even data gathered  from simple wearable devices  would allow real investigation of the diurnal patterns of the skin and link them to  
stress, sun exposure, air pollution, food consumption, cosmetic usage and many more features of daily life.
This would help cosmetic research organisations  to add an extra layer of information on compliance of volunteers as part of regular clinical trials
Without the need of having volunteers coming on a daily basis to a clinic facility.

\subsection*{Acknowledgements}
Funding for Alan Smeaton's work in this paper was provided by Science Foundation Ireland under grant number SFI/12/RC/2289\_P2 and equipment and lab expertise was provided by Oriflame R\&D.

\section*{CONFLICTS OF INTEREST}
The authors state no conflicts of interest.

\section*{Supporting Information}

The data used in this research including images and analysis is archived in a public repository and is available for private reviewer viewing at \url{https://figshare.com/s/4d95ba3def3ea3b3f87c} but which is not to be referenced.  The DOI for this will be \url{10.6084/m9.figshare.11881059} when the dataset is published depending on the outcome of this paper submission.

\bibliographystyle{plain} 
\bibliography{WileyNJD-Doc}%

\end{document}